\documentclass[letterpaper]{article} 
\usepackage{aaai24}  
\usepackage{times}  
\usepackage{helvet}  
\usepackage{courier}  
\usepackage[hyphens]{url}  
\usepackage{graphicx} 
\usepackage{natbib}  
\usepackage{caption} 
\usepackage{algorithm}
\usepackage{algorithmic}
\usepackage{booktabs}       
\usepackage{amsfonts}       
\usepackage{nicefrac}       
\usepackage{microtype}      
\usepackage{xcolor}         

\usepackage{multirow}
\usepackage[caption=false,font=footnotesize]{subfig}
\usepackage{makecell}
\usepackage{amsmath}
\usepackage{amssymb}
\usepackage{bm}
\usepackage{rotating}

\usepackage{makecell}
\usepackage{tabularx}
\usepackage{bbding}
\usepackage{newfloat}
\usepackage{listings}
\DeclareCaptionStyle{ruled}{labelfont=normalfont,labelsep=colon,strut=off} 
\floatstyle{ruled}
\newfloat{listing}{tb}{lst}{}
\floatname{listing}{Listing}
\title{Signed Graph Neural Ordinary Differential Equation for Modeling Continuous-time Dynamics}
\author{
    Lanlan Chen\textsuperscript{\rm 1},
    Kai Wu\textsuperscript{\rm 2}\footnote{Corresponding author},
    Jian Lou\textsuperscript{\rm 3},
    Jing Liu\textsuperscript{\rm 1}
}
\affiliations{
    \textsuperscript{\rm 1}Guangzhou Institute of Technology, Xidian University\\
    \textsuperscript{\rm 2}School of Artificial Intelligence, Xidian University\\
    \textsuperscript{\rm 3}Zhejiang University\\

    21181214030@stu.xidian.edu.cn,
    kwu@xidian.edu.cn,
    jian.lou@hoiying.net,
    neouma@163.com
}

\usepackage{bibentry}

\begin{document}

\maketitle

\begin{abstract}
Modeling continuous-time dynamics constitutes a foundational challenge, and uncovering inter-component correlations within complex systems holds promise for enhancing the efficacy of dynamic modeling. The prevailing approach of integrating graph neural networks with ordinary differential equations has demonstrated promising performance. However, they disregard the crucial signed information intrinsic to graphs, impeding their capacity to accurately capture real-world phenomena and leading to subpar outcomes.

In response, we introduce a novel approach: a signed graph neural ordinary differential equation, adeptly addressing the limitations of miscapturing signed information. Our proposed solution boasts both flexibility and efficiency. To substantiate its effectiveness, we seamlessly integrate our devised strategies into three preeminent graph-based dynamic modeling frameworks: graph neural ordinary differential equations, graph neural controlled differential equations, and graph recurrent neural networks. Rigorous assessments encompass three intricate dynamic scenarios from physics and biology, as well as scrutiny across four authentic real-world traffic datasets. Remarkably outperforming the trio of baselines, empirical results underscore the substantial performance enhancements facilitated by our proposed approach.Our code can be found at https://github.com/beautyonce/SGODE.
\end{abstract}

\section{Introduction}

\begin{table*}[!htbp]
	\centering
        \footnotesize
	\begin{tabularx}{\textwidth}{lXXl}
		\toprule
		 \textbf{Method} & \textbf{Formula} & \textbf{Graph Information} & \textbf{Signed ?} \\
		\midrule
		 NDCN \cite{Zang2020Neural} 
		& $\frac{d\boldsymbol{H}(t)}{dt}=f\left(\boldsymbol{\widetilde{A}}, \boldsymbol{H}(t), \boldsymbol{\theta}\right)$  
  & $\boldsymbol{\widetilde{A}}=\boldsymbol{D}^{-\frac{1}{2}}(\boldsymbol{D}-\boldsymbol{A})\boldsymbol{D}^{-\frac{1}{2}}$ 
  & \XSolidBrush\\
 STGODE \cite{Fang2021Spatial} & 
 $\frac{d\boldsymbol{H}(t)}{dt}=f(\boldsymbol{\widetilde{A}}, \boldsymbol{H}(t), \boldsymbol{H}(0), \boldsymbol{\theta})$
 & $\boldsymbol{\widetilde{A}}={\frac{\alpha}{2}(\boldsymbol{I}+\boldsymbol{D}}^{-\frac{1}{2}}\boldsymbol{A}\boldsymbol{D}^{-\frac{1}{2}})$ 
 & \XSolidBrush \\
 STG-NCDE \cite{choi2022graph} &
 $\frac{d\boldsymbol{H}\left(t\right)}{dt}=f(\boldsymbol{\widetilde{A}}, \boldsymbol{H}(t), \boldsymbol{\theta})$
 & $\boldsymbol{\widetilde{A}}=\boldsymbol{I}+\phi\left(\sigma\left(\boldsymbol{E}\cdot \boldsymbol{E}^T\right)\right)$
 & \XSolidBrush \\
SGODE (This Work) & $\frac{d\boldsymbol{H}(t)}{dt}=f\left(\boldsymbol{K}, \boldsymbol{H}(t),\boldsymbol{B}(t),\boldsymbol{\theta}\right)$
& $\boldsymbol{K}=\sigma(\boldsymbol{E}_{1}\cdot \boldsymbol{E}_{2}^T)+(-\sigma(\boldsymbol{E}_{3}\cdot \boldsymbol{E}_{4}^T))$
& \Checkmark\\
\bottomrule
\end{tabularx}
\caption{ODE-GNN approaches. \(\boldsymbol{A}\in \mathbb{R}^{n\times n}\) is the adjacency matrix of the network, and \(\boldsymbol{D}\in \mathbb{R}^{n\times n}\) is the corresponding degree matrix. \(\phi\) is a softmax activation, and \(\sigma\) is a rectified linear unit. \(\boldsymbol{I}\in \mathbb{R}^{n\times n}\) is the identity matrix. \(\boldsymbol{E}\in \mathbb{R}^{n\times d}\) represents the node embedding matrix, where \(d\) is the embedding dimension, much smaller than \(n\).
\(\boldsymbol{H}\left(t\right)\) represents the state feature in the hidden space at time \(t\). In this context, $\boldsymbol{\theta}$ refers to the other parameters. While other methods utilize the normalized adjacency matrix \(\boldsymbol{\widetilde{A}}\in \mathbb{R}^{n\times n}\) of a homogeneous graph, this study constructs a signed coefficient matrix \(\boldsymbol{K}\in \mathbb{R}^{n\times n}\) composed of positive and negative coefficient matrices, which is crucial for capturing different types of information.}
\label{tab:GNN-ODE}
\vskip -0.2in
\end{table*}
Complex systems prevalent in the real world, such as gene regulation \cite{marbach2012wisdom}, social networks \cite{wasserman1994social}, climate models \cite{Hwang2021Climate}, and traffic systems \cite{zhao2019t}, often find representation as complex networks governed by nonlinear dynamics \cite{lieberman2005evolutionary}. 
In contrast to deterministic and easily obtainable graphs \cite{wu2020comprehensive, feng2023exploring}, the graph structures of these complex networks are challenging to explicitly articulate.
Despite the extensive exploration of nonlinear dynamical systems, a significant number of complex networks  continue to evade a clear understanding of their underlying dynamics.

In recent years, a noteworthy trend has emerged, involving the fusion of ordinary differential equations (ODEs) with neural networks to acquire insights into continuous-time dynamics 
{\cite{NEURIPS2018_NODE,NEURIPS2019_LatentNODE,kidger2020neural,jhin2021ace,Hwang2021Climate,huang2021coupled,Fang2021Spatial,choi2022graph,Jhin2022EXIT}.
These hybrids, 
encompassing  both ODEs and graph neural networks (GNNs), have demonstrated promising performance across a variety of tasks. This includes climate modeling \cite{Hwang2021Climate,jhin2021ace}, traffic flow prediction \cite{poli2019graph,Fang2021Spatial,choi2022graph}, node classification \cite{Zang2020Neural,xhonneux2020continuous}, and dynamic interactions \cite{huang2021coupled}. 
However, it remains pertinent to note that a substantial portion of complex networks retains enigmatic dynamics yet to be fully unraveled.

The majority of existing methodologies predominantly focus on either inferring or utilizing unsigned graphs, wherein only the presence or absence of dependencies between nodes is taken into account, while the type of edges are disregarded. As shown in Table \ref{tab:GNN-ODE}, the current GNN-ODE methods are unable to capture and use signed information of dynamics, thus rendering this problem challenging. It is noteworthy that within unsigned graphs, nodes face challenges in shifting towards the opposing trend (rise or decline) when they and their neighboring nodes align in a similar variation trend (decline or rise). This constraint inherently restricts the capacity to represent dynamic processes effectively.
A remedy for this limitation is the introduction of signed connections, which can markedly enhance the scenario. Intriguingly, signed graphs find applicability across a multitude of complex systems \cite{shi2019dynamics}. Notable instances include predation and prey relationships within ecosystems, activation and repression dynamics in gene regulation networks \cite{karlebach2008modelling}, as well as cooperation and antagonism dynamics observed within social and economic networks \cite{derr2018signed}. To offer a concrete illustration, we examine the context of a three-way intersection within traffic systems, thereby illustrating the presence of signed graphs (refer to Appendix).

In the context of an unknown underlying graph, one approach involves inferring the graph structure, which is inherently tied to the realm of graph structure learning. However, current graph structure learning methods fall short in handling signed information. 
To address this challenge, various strategies have been explored. One approach entails capturing the edge distribution \cite{kipf2018neural,franceschi2019learning,shang2021discrete}, followed by sampling the graph's adjacency matrix from this distribution. Yet, the complexity arises when assuming a bipartite adjacency matrix and attempting separate learning of its positive and negative edges. This introduces intricate optimization issues due to the inherent uncertainty tied to the sampled graph. Additionally, distinct optimization processes for positive and negative graphs exacerbate uncertainties, impeding the achievement of convergence. Moreover, approximating the edge distribution through neural networks imposes substantial computational and storage demands on a high-precision ODE solver. Alternatively, an approach directly embeds the graph into a matrix composed of learnable parameters within a GNN 
\cite{GraphWaveNet, NEURIPS2020_Adaptive, wu2020connecting, deng2021graph, choi2022graph, DNODE, jiang2023spatio}, enabling simultaneous optimization alongside other pertinent parameters. While methods employing learnable parameter matrices are simple to optimize, they often neglect signed information and remain susceptible to overfitting. Hence, there is a compelling need to devise a GNN-ODE method that strikes a balance between ease of optimization, consideration of signed information, preservation of stability, and mitigation of overfitting.

We propose a \textbf{Signed Graph Neural Ordinary Differential Equation}, called SGODE, to capture and use positive and negative information of nodes during continuous dynamics. Our contributions are summarized as follows: 

1) The significance of signed information in real-world scenarios is undeniable. In response, we propose a straightforward yet impactful solution that effectively addresses the limitations of the current GNN-ODE methodologies by capturing and leveraging this crucial information.

2) SGODE offers a high degree of flexibility and seamlessly integrates into various frameworks for graph-based dynamic modeling, including graph neural ODEs, graph neural controlled differential equations (NCDEs), and graph recurrent neural networks (RNNs). Empirical results across multiple dynamic modeling datasets and traffic flow datasets substantiate the effectiveness of SGODE.

\section{Background}

\textbf{Notations}. We denote the training data containing $n$ time series as $\boldsymbol{X}$. $\boldsymbol{X}^i$ represents the features of the $i$-th time series, and $\boldsymbol{X}(t)$ encompasses the features of all nodes at time $t$. The remaining notations are described in Table \ref{tab:GNN-ODE}.
Specifically, if the relationship between node $i$ and node $j$ is positive, we have $\boldsymbol{K}_{ij}>0$, while a negative relationship corresponds to $\boldsymbol{K}_{ij}<0$. An edge is absent when $\boldsymbol{K}_{ij}=0$.

For synthetic dynamics, there are a total of $S$ time steps for training, written as $\boldsymbol{X}_S=\{\boldsymbol{X}_{s1}, \boldsymbol{X}_{s_2}, \dots, \boldsymbol{X}_{s_s} \}$, and $P$ steps for forecasting, written as $\boldsymbol{X}_P=\{\boldsymbol{X}_{p_1}, \boldsymbol{X}_{p_2}, \dots, \boldsymbol{X}_{p_p}\}$. If the time step intervals in $S$ and $P$ are fixed, the sampling is considered as equal intervals sampling; otherwise, it is referred to as irregular sampling. Given $S, \boldsymbol{X}_S$ and a set of times $P$, the model needs to predict $\boldsymbol{X}_P$. By using $\boldsymbol{\theta}$ to represent other parameters in the network except $\boldsymbol{K}$, the model is denoted as $\boldsymbol{\widehat X}_P=g(\boldsymbol{K}, \boldsymbol{\theta}, \boldsymbol{X}_S)$.
For traffic flow data, we utilize a T-step window approach to forecast the subsequent $\tau$ steps.
If $t+1$ represents the initial time step of a certain window, the model can be denoted as $\boldsymbol{\widehat X}_{t+T+1:t+T+\tau}=M(\boldsymbol{K}, \boldsymbol{\theta}, \boldsymbol{X}_{t+ 1:t+T})$. Take $\mathcal L$ to represent the loss function of the predicted value and ground truth. Then our goal is $\mathop{\arg \min}_{\boldsymbol{K}, \boldsymbol{\theta}} {\textstyle \sum_t}\mathcal L(M(\boldsymbol{K}, \boldsymbol{\theta},\boldsymbol{X}_{z_1}),\boldsymbol{X}_{z_2})$
For synthetic dynamics, we set $z_1=S$, $z_2=P$, and for traffic flow prediction, we set $z_1=t+1:t+T$, $z_2=t+T+1:t+T+\tau$ for all the training examples that are partitioned by window.  In the following, we review of the relevant models, including NDCN \cite{Zang2020Neural}, STG-NCDE \cite{choi2022graph}, and DCRNN \cite{li2018diffusion}.

\textbf{NDCN}. NDCN \cite{Zang2020Neural} can be considered as either a continuous-time GNN or a graph neural ODE model. NDCN utilizes an encoding function to transform $\boldsymbol{X}(t)$ into the hidden space, employs a continuous model to regulate the dynamics on the graph within the hidden space, and subsequently decodes the hidden state back into the original space. For irregular and  equal intervals sampling tasks, NDCN uses $L_1$ loss. The ODE layer is
\begin{align}
\frac{d \boldsymbol{H}(t)}{d t} =\sigma \left(\boldsymbol{\widetilde{A}} \boldsymbol{H}(t) \boldsymbol{W}_h+\boldsymbol{b}_h\right),\label{ndcn} 
\end{align}
where $\boldsymbol{W}_h\in \mathbb{R}^{n\times h}$ and $\boldsymbol{b}_h\in \mathbb{R}^{h}$ are the parameters of fully connected layer $\mathrm{FC}$. NDCN adopts a linear diffusion operator $\boldsymbol{\widetilde{A}}$ (shown in Table \ref{tab:GNN-ODE}).

\textbf{STG-NCDE}. STG-NCDE \cite{choi2022graph} STG-NCDE \cite{choi2022graph} combines the advantages of graph convolutional network and NCDE to design a unified space-time NCDE framework, which encompasses two NCDEs, a CDE that generates a continuous path for each node, denoted as $f$, and another CDE applies a method for spatial and temporal processing jointly, denoted as $g$. Denote the hidden states of $f$ and $g$ as $\boldsymbol{H}(t)$ and $\boldsymbol{Z}(t)$ respectively, then we have
\begin{equation}
    \frac{d}{dt} \begin{bmatrix}
 \boldsymbol{Z}(t)\\
\boldsymbol{H}(t)
\end{bmatrix}=\begin{bmatrix}
 g(\boldsymbol{Z}(t);\boldsymbol{\theta}_g)f(\boldsymbol{H}(t);\boldsymbol{\theta}_f)\frac{d\boldsymbol{X}(t)}{dt}  \\
f(\boldsymbol{H}(t);\boldsymbol{\theta}_f)\frac{d\boldsymbol{X}(t)}{dt}
\end{bmatrix}.
\end{equation}
The initial value is determined by the fully connected layer: $\boldsymbol{H}(0)=\mathrm{FC}_{dim(\boldsymbol{X}^i)\rightarrow dim(\boldsymbol{H}^i)}(\boldsymbol{X}({t_0}))$, $\boldsymbol{Z}(0)=\mathrm{FC}_{dim(\boldsymbol{H}^i)\rightarrow dim(\boldsymbol{Z}^i)}(\boldsymbol{H}(0))$.  $\boldsymbol{\theta}_{f}$ is the parameters of the CDE function $f$. $\mathrm{FC}_{input\_size\rightarrow output\_size}$ means a fully-connected layer whose input size is $input\_size$ and output size is
$output\_size$. $\theta_f$ is the parameters of the CDE function $f$. $f$ is composed of $l$ layers of MLPs, with the activation function in the final layer being $\psi$ (hyperbolic tangent). The activation functions in the other layers are \(\sigma\). \(g\) and \(f\) share similar structures, the only difference being that \(g\) incorporates adaptive graph information in the middle layer. Let $\boldsymbol{Z}_{B_{0}}(t)$ and $\boldsymbol{Z}_{B_{1}}(t)$ be the input and output of this layer. The equation is given by
\begin{equation}
\boldsymbol{Z}_{B_{1}}(t) =(\boldsymbol{I}+\phi(\sigma(\boldsymbol{E\cdot E^T})))\boldsymbol{Z}_{B_{0}}(t)\boldsymbol{W}_{s},\label{ncde-g-adap}
\end{equation}
where $\boldsymbol{I}$ is the $n \times n$ identity matrix. $\boldsymbol{E}$ is a trainable node embedding matrix, $\boldsymbol{E}^T$ is its transpose.  Eq. \ref{ncde-g-adap}  has two important constituent components $\boldsymbol{I}+\phi(\sigma(\boldsymbol{E\cdot E^T}))$ and $\boldsymbol{W}_{s}$
\cite{NEURIPS2020_Adaptive}, the former adaptively learns spatial dependencies, and the latter extracts the specific pattern of each node through $\boldsymbol{W}_{s}=\boldsymbol{E}\cdot$$\boldsymbol{W}_{pool}$, where $\boldsymbol{ E}\in \mathbb{R}^{n\times d}$ and $\boldsymbol{W}_{pool}\in \mathbb{R}^{d\times h\times h }$ is weight pool for $\boldsymbol{E}$.

\textbf{DCRNN}. DCRNN \cite{li2018diffusion} introduces diffusion graph convolutions to capture spatial dependencies, facilitating the modeling of traffic flow dynamics from a spatiotemporal perspective. 
Within DCGRU, GRU \cite{chung2014empirical} serves as a recurrent neural network to simulate time dependence, with matrix multiplication being replaced by diffusion convolution,
\begin{equation}
\label{eqGRU}
\begin{split}
   &\boldsymbol{R}({t})=\operatorname{sigmoid}\left(\boldsymbol{W}_{\boldsymbol{R}} \star_{A}\left[\boldsymbol{X}({t}) \| \boldsymbol{H}({t-1})\right]+b_{R}\right), \\
   &\boldsymbol{C}({t})=\tanh \left(\boldsymbol{W}_{\boldsymbol{C}} 
   \star_{\boldsymbol{A}}\left[X({t}) \|\left(\boldsymbol{R}({t}) \odot \boldsymbol{H}({t-1})\right]+b_{C}\right),\right. \\
   &\boldsymbol{U}({t})=\operatorname{sigmoid}\left(\boldsymbol{W}_{U} \star_{A}\left[\boldsymbol{X}({t}) \| \boldsymbol{H}({t-1})\right]+b_{U}\right),\\
   &\boldsymbol{H}({t})=\boldsymbol{U}({t}) \odot \boldsymbol{H}({t-1})+\left(1-\boldsymbol{U}({t})\right) \odot \boldsymbol{C}({t}),
\end{split}
\end{equation}
where diffusion convolution $\star_{\boldsymbol{A}}$ is defined as,
\begin{equation}
\begin{aligned}
    \label{eq:GConv}
    \boldsymbol{W}_{Q} \star_{\boldsymbol{A}} &\boldsymbol{H}(t)=\\
    &\sum_{m}\left(w_{m_1}^{Q}\left(\boldsymbol{D}_{O}^{-1} \boldsymbol{A}\right)^{m}+w_{m_2}^{Q}\left(\boldsymbol{D}_{I}^{-1} \boldsymbol{A}^{T}\right)^{m}\right) \boldsymbol{H}(t). \nonumber
    \end{aligned}
\end{equation}
$\boldsymbol{D}_O$ and $\boldsymbol{D}_I$ are out-degree and in-degree matrix, $||$ is connected along the feature dimension, and $\odot$ denotes element-wise product. The diffusion step $m$ is a hyperparameter, $w_{m_1}^{Q}, w_{m_2}^{Q}, b_Q$ for $Q =R, U, C$ are all model parameters. For simplicity, our expression here follows the conventions of GTS \cite{shang2021discrete}.

\section{Methods}
\textbf{Signed Graph Ordinary Differential Equation}.
Learning the distribution of edges in a signed graph is a challenging optimization task, particularly when the edge distribution must be learned through a large neural network. When the number of nodes expands to hundreds, high-precision ODE solvers encounter difficulties in calculation and storage. Therefore, we opt for direct initialization of the corresponding node vector for each node. We consider two forms of coefficient matrix $\boldsymbol{K}$, namely $\boldsymbol{K}=\boldsymbol{E}_1\boldsymbol{E}_2^T$ and $\boldsymbol{K}=\boldsymbol{K}_{pos}+\boldsymbol{K}_{neg}$, where $\boldsymbol{K}_{pos}=\sigma(\boldsymbol{E}_{1} \boldsymbol{E}_{2}^T), \boldsymbol{K}_{neg}=-\sigma(\boldsymbol{E}_{3}\boldsymbol{E}_{4}^T)$, $\boldsymbol{E}_{*}\in\mathbb{R}^{n\times d}$. The first form indicates learning the positive and negative relations of each node pair. The second form indicates that only the positive and negative relations of some node pairs are learned, and the node pairs with low dependency are ignored. We do not use any other constraints to ensure that $\boldsymbol{K}$ learns the connection weights. 

Without constraints, the evolution of unkonwn signed graph ODEs is unstable due to the extensive parameter space and the random initialization of embedding vectors. Motivated by the integration of initial value-related terms into the ordinary differential equation for neighborhood information learning while retaining original features \cite{xhonneux2020continuous}, we introduce self-trend features $\boldsymbol{B}^i$ associated with the state of the $i$th node, augmenting the stability of the dynamics learning process. The inclusion of $\boldsymbol{B}(t)$ implies that the instantaneous rate of change in node features is influenced not just by interactions with other nodes, but also by its inherent changes. Learning self-evolving features is comparably straightforward compared to graph-related learning processes, allowing for representation of temporal variations.
We posit that $\boldsymbol{B}(t)$ primarily encompasses three factors: the constant term, initial features, and the features at the current time step. Formally, the interaction of information on signed graphs is defined as follows:
\begin{equation}
\label{eq:SGODE}
    \frac{\mathrm{d} \boldsymbol{H}({t})}{\mathrm{d} t}=f(\boldsymbol{K}\boldsymbol{H}(t)\boldsymbol{W}_h+\boldsymbol{B}(t)),
\end{equation}
\begin{equation}
\label{eq:B(t)}
\boldsymbol{B}(t)=\lambda_1g_1(\boldsymbol{H}(t))+\lambda_2g_2(\boldsymbol{H}(0))+\lambda_3\boldsymbol{B}_0,
\end{equation}
where both \(f\) and \(g\) simply refer to functions, and $\boldsymbol{B}_t$ describes the information of the state trend at time $t$. $\lambda_1, \lambda_2, \lambda_3 \in [0,1]$ indicates whether the item is taken. 
We emphasize the significance of focusing on current time-step features, whether for longer-term predictions or shorter periods within the prediction window (e.g., \(\lambda_1 = 1\)). Even incorporating basic linear relationships can effectively enhance the performance of the ODE model.

\textbf{Embedding SGODE into NDCN}. 
We first embedding SGODE into NDCN \cite{Zang2020Neural} to show its effectiveness of modeling continuous-time dynamics.
We replace the intermediate ODE layer (Eq. \ref{ndcn}) with our proposed SGODE layer, as described in Eq. \ref{eq:SGODE},
\begin{equation}
\frac{d \boldsymbol{H}(t)}{d t} =\sigma \left(\boldsymbol{K} \boldsymbol{H}({t}) \boldsymbol{W}_h+\boldsymbol{B}({t})+\boldsymbol{b}_h\right).
\end{equation}
We consider three forms of $\boldsymbol{K}$. The first form directly represents $\boldsymbol{K}$ by $n \times n$ learnable parameters, called $\textbf{SGODEv1}$. The second is to express $\boldsymbol{K}=\boldsymbol{E}_1\boldsymbol{E}_2^{T}$ using the similarity calculated by two node embedding vectors, denoted as $\textbf{SGODEv2}$. 
The third is to utilize two node embedding vectors to calculate the positive relationships, and employ two node embedding vectors to calculate the negative relationships, $\boldsymbol{K}={\sigma}(\boldsymbol{E}_1\boldsymbol{E}_2^{T})-{\rm \sigma}(\boldsymbol{E}_3\boldsymbol{E}_4^ {T})$, denoted as \textbf{SGODEv3}. Given the simplicity of the synthetic dynamics, in \textbf{SGODEv3}, we establish a straightforward linear scaling relationship between $\boldsymbol{B}({t})$ and $\boldsymbol{H}({t})$, as follows: $\boldsymbol{B}({t})=\boldsymbol{b}\boldsymbol{H}({t})\boldsymbol{W}_h$, where $\boldsymbol{b}\in \mathbb{R}^{n}$. In practice, these learnable parameters are trained jointly with other network parameters. Further details can be found in Appendix.

\textbf{Embedding SGODE into STG-NCDE}. 
SGODE-NCDE refer to the STG-NCDE variant by modify the adaptive graph convolution layer of Eq.\ref{ncde-g-adap} with the proposed SGODE. We observe that $\boldsymbol{E}$ for constructing the adaptive adjacency matrix is closely related to $\boldsymbol{W}_{S}$ for extracting patterns of each node. In order to establish a relevant connection between $\boldsymbol{K}$ and $\boldsymbol{W}_{S_1}$, we adopt a node embedding vector $\boldsymbol{E}_1$ 
that controls both positive and negtive relations and get weight pool of each node with $\boldsymbol{W}_{S_1}$.
To enhance the sparsity of the graph, we employ an adaptive mask matrix \cite{Jhin2021Attentive}. Formally,
\begin{equation}
\begin{aligned}
    \boldsymbol{K}_0={\sigma}(\boldsymbol{E}_1\boldsymbol{E}_2^{T})-{\sigma}(\boldsymbol{E}_1\boldsymbol{E}_3^ {T}), \\\boldsymbol{W}_{s_1}=\boldsymbol{E}_1 \cdot \boldsymbol{W}_{pool_1},
    \boldsymbol{W}_{s_2}=\boldsymbol{E}_1 \cdot \boldsymbol{W}_{pool_2},
\end{aligned}
\end{equation}
and construction of the adaptive mask matrix follows these steps: initially, a learnable matrix is set as $\boldsymbol{M}=\boldsymbol{E}_{M1}\cdot \boldsymbol{E}_{M2}^T$. Subsequently, a hard sigmoid function is employed for rigorous classification, and the result is rounded up to yield the final mask matrix, denoted as $\boldsymbol{M}$. The detailed training approach is outlined below:
\begin{equation}
\varphi(x)=round (\operatorname{hardsigmoid}(\alpha x)),
\end{equation}
for the forward path, and
\begin{equation}
\nabla \varphi(x)=\nabla \operatorname{hardsigmoid}(\alpha x), 
\end{equation}
for the backward path, where the temperature $\alpha \ge 1.0$ is a hyperparameter to control the slope of the sigmoid function. Given the presence of $\boldsymbol{W}_{s_1}$ for interaction information extraction. We then define $g_1(\boldsymbol{Z}_{B_0}(t))=\boldsymbol{Z}_{B_0}(t)\cdot \boldsymbol{W}_{s_2}=\boldsymbol{Z}_{B_0}(t) \boldsymbol{E}_1\boldsymbol{W}_{pool_2}$.
Finally, we have
$\boldsymbol{K}=\varphi(\boldsymbol{M}) \odot \boldsymbol{K}_0$, and 
    $\boldsymbol{B}(t)  = \boldsymbol{Z}_{B_0}(t)\cdot\boldsymbol{W}_{s_2}+\boldsymbol{B}_0$.
i.e. we substitute this equation for Eq. \ref{ncde-g-adap}.

\textbf{Embedding SGODE into DCRNN}. 
SGODE-RNN refers to the DCRNN variant.
Our continuous graph diffusion method,
\begin{equation}
    \label{eq:ode-rnn1}
    \boldsymbol{W}_{\boldsymbol{Q}} \star_{\boldsymbol{K}} \boldsymbol{H}(t)=\sum_{m}\boldsymbol{w}_{t_m}^{Q}\boldsymbol{H}({t_m}),  
\end{equation}
\begin{equation} \label{eq:ode-rnn2}
\frac{\mathrm{d} \boldsymbol{H}({t_m})}{\mathrm{d} t}=
\boldsymbol{K}\boldsymbol{H}(t_m)\boldsymbol{W}_h+\boldsymbol{B}(t_m),
\end{equation}
depicts the continuous dynamics of hidden states on a signed graph during state transitions and the extraction of information on continuous dynamics. We establish a two-layer FC network as $g_2(\boldsymbol{H}(0))=\mathrm{FC}_{\operatorname{dim} (\boldsymbol{H}^i) \rightarrow \operatorname{dim}(\boldsymbol{H}^i)}(\sigma(\mathrm{FC}_{\operatorname{dim} (\boldsymbol{H}^i) \rightarrow \operatorname{dim}(\boldsymbol{H}^i)}(\boldsymbol{H}(t))))$, and  
$g_1(\boldsymbol{H}(t_m))=\boldsymbol{b}\boldsymbol{H}(t)\boldsymbol{W}_h$.
Here, $\boldsymbol{K}={\sigma}(\boldsymbol{E}_1\boldsymbol{E}_2^{T})-{\sigma}(\boldsymbol{E}_3\boldsymbol{E}_4^ {T})$. The ODE within the interval $[t,t+1]$ can be perceived as a normalization of an ODE of arbitrary length. To enhance monitoring of the internal evolution within the ODE solver and mitigate error accumulation, we extract features from a set of $m$ equidistantly sampled states within the range of $[0,1]$. If $m=1$, we extract only the starting and ending information of ODE.
This strategy enables an increased focus on the inherent trend of self-evolution, while also taking into account node interactions and self-evolution features, thereby enhancing the model's fitting capabilities. 
\subsection{Training Loss}
The training loss function in our model is shown as follows,
\begin{equation}
    \label{loss}
    \mathcal L=\frac{1}{|P|}\frac{1}{|\tau|} {\textstyle \sum_{p\in P}}{\textstyle \sum_{\tau}|\boldsymbol{\widehat X}({P(t)})-\boldsymbol{X}(P(t))|},
\end{equation}
$p$ refers to each training example in the training set $P$, $\tau$ refers to the time step to be predicted in each training example, $|P|$ is the number of training examples, and $|\tau|$ is the prediction step number.
\section{Experiments}

\subsection{Experimental Setup}
\subsubsection{Dataset}\textit{Three Continuous-time Dynamics}. We follow the dataset settings in NDCN \cite{Zang2020Neural}. We generate three continuous-time dynamics: heat diffusion, mutualistic interaction, and gene regulation dynamics on five graphs, including 1) Grid network, where each node is connected with 8 neighbors; 2) Random network \cite{erdos1959random}; 3) Power-law network \cite{barabasi1999emergence}; 4) Small-world network \cite{watts1998collective}; 5) Community network \cite{fortunato2010community}. We categorize irregular sampling into two types: interpolated value prediction and extrapolated value prediction. The training, interpolation prediction, and extrapolation prediction are allocated in a ratio of 80/20/20, respectively. The details of three continuous-time dynamics and data generation are shown in Appendix.

\textit{Traffic Prediction Dataset}.
Four publicly available real-world traffic datasets are employed: (1) METR-LA and PEMS-BAY \cite{li2018diffusion}. METR-LA contains the traffic information on the highways of Los Angeles County and consists of 207 sensors, collecting data over a span of 4 months. PEMS-BAY comprises 325 sensors in the Bay Area, with data collected over a period of 6 months. (2) PeMS traffic datasets \cite{chen2001freeway}, previously employed in other works \cite{Fang2021Spatial,choi2022graph}, include PeMSD4 and PeMSD8 with 307 and 170 nodes, respectively. The frequency of data in these four datasets is uniformly set to 5 minutes. For METR-LA and PEMS-BAY, we adopt the train/valid/test split of 70\%/10\%/20\% as suggested by GTS \cite{shang2021discrete}, while for PeMSD4 and PeMSD8, we follow STG-NCDE \cite{choi2022graph} and use a split of 60\%/20\%/20\%.

\paragraph{Baselines}
To show the effectiveness of SGODE on modeling continuous dynamics, the following basedlines are employed. (1) \textbf{NDCN} \cite{Zang2020Neural}. It is a representative algorithm for ODE-GNN methods using real graphs. (2) \textbf{No-graph}. We remove graph structure in NDCN as the benchmark algorithm directly. (3) \textbf{Adaptive-NDCN}. For a fair comparison, we replace the linear map in NDCN with a learnable matrix $\phi(\sigma(E_1E_2^T))$ \cite{NEURIPS2020_Adaptive,choi2022graph,GraphWaveNet,wu2020connecting}. (4) \textbf{GTS-NDCN}. For a fair comparison, we replace the linear map in NDCN with a learnable matrix proposed in \cite{shang2021discrete}. The details of \textbf{SGODEv1}, \textbf{SGODEv2}, and \textbf{SGODEv3} are shown in the \textbf{Methods} section.

For traffic prediction problems, We compare with widely used time series regression models, including (1) \textbf{HA}: Predicting based on the historical average; (2) \textbf{ARIMA}: Forecasting based on the statistical characteristics of stationary time series.
(3) \textbf{VAR}: Vector Auto-Regression. (4) \textbf{SVR}: Support Vector Regression which uses linear support vector machine for the regression task; The following deep neural network based approaches are also included: (5) \textbf{FNN}: Feed-forward Neural Network ; (6) \textbf{LSTM}: Recurrent Neural Network with fully connected LSTM hidden units \cite{sutskever2014sequence}; (7) \textbf{DCRNN} \cite{li2018diffusion}: a deep
learning framework for traffic forecasting that incorporates both spatial and temporal dependency in the traffic flow; (8) \textbf{LDS} \cite{franceschi2019learning}: the model considers graphs as hyperparameters within a two-layer optimization framework, where it learns parameterized element-wise Bernoulli distributions; (9) \textbf{GraphWaveNet} \cite{GraphWaveNet}: the most representative deep models for traffic forecasting; (10) \textbf{MTGNN} \cite{wu2020connecting} is
an extended version of GraphWaveNet that extends the adaptive
graph leaning part; (11) \textbf{GTSv} and \textbf{GTS} \cite{shang2021discrete}: they are variants that apply inference graphs to T-GCN \cite{zhao2019t} and DCRNN models respectively. (12) \textbf{STEP} \cite{shao2022pre} is
an extension of spatial-temporal GNN enhanced by a scalable time series Pre-training model; (13) \textbf{MegaCRN} \cite{jiang2023spatio}: they designed meta-graph learner for spatiotemporal graph learning, to explicitly disentangles
the heterogeneity in space and time.
To additionally demonstrate the performance of SGODE embedded in the existing NCDE framework, we compare it with the advanced STGODE \cite{Fang2021Spatial} and STG-NCDE \cite{choi2022graph}.

\paragraph{Hyperparameters.}
we set the learning rate to 0.005 for SGODE-RNN and to 0.001 for SGODE-NCDE. 
We have reproduced the
results for four methods, i.e., NDCN, GTS, STG-NCDE, and MegaCRN. If the original paper provides hyperparameters
(e.g., STG-NCDE, MegaCRN), we adhere to the same settings there. If not (e.g., GTS), we conduct experiments within
their recommended hyperparameter ranges. Specifically, for GTS, we perform a grid search on hyperparameters including
learning rate, number of clusters, and regularization weight.
For MegaCRN, we set the number of meta-nodes to 20. Detailed parameter settings for SGODE and baselines are available in Appendix.

These methods are evaluated based on three commonly used metrics, including (1) Mean Absolute Error (MAE), (2) Mean Absolute Percentage Error (MAPE), and (3) Root Mean Squared Error (RMSE). To maintain consistency with the referenced algorithms, we refer to DCRNN \cite{li2018diffusion} for the evaluation index calculation method on METR-LA and PEMS-BAY, and refer to STG-NCDE \cite{choi2022graph} on PeMS04 and PeMSD8. 
\begin{table}[htb]
\vskip -0.1in
  \centering
  \resizebox{\linewidth}{!}{
    \begin{tabular}{crrrrrr}
    \toprule
          &   Methods    & \multicolumn{1}{l}{Grid} & \multicolumn{1}{l}{Random} & \multicolumn{1}{l}{Power} & \multicolumn{1}{l}{Small} & \multicolumn{1}{l}{Com.} \\
    \midrule
    \multirow{5}[12]{*}{} & No-graph & 41.1±0.1 & 10.1±16.8 & 20.7±0.1 & 21.2±0.2 & 24.2±2.5 \\
          & NDCN  & \underline{3.3±0.4} & 3.4±0.7 & 6.0±0.1 & 3.6±0.1 & 3.7±0.3 \\
     I     & Adaptive-NDCN & 7.9±2.0 & 8.3±2.8 & 12.6±4.2 & 9.8±2.2 & 9.3±3.0 \\
     & GTS-NDCN & 9.9±2.1 & 5.0±1.1 & 7.4±1.2 & 8.5±1.6 & 10.0±3.5 \\
       \cmidrule{2-7}  & \textbf{SGODEv1} & 4.5±1.3 & \textbf{1.5±0.4} & 3.2±0.5 & 4.7±0.7 & \underline{2.1±0.6} \\
          & \textbf{SGODEv2} & \textbf{2.5±0.9} & \underline{2.0±0.5} & \textbf{1.7±0.5} & \textbf{1.7±0.4} & \textbf{1.7±0.6} \\
          & \textbf{SGODEv3} & 3.4±1.3 & \underline{2.0±0.8} & \underline{2.8±0.4} & \underline{2.9±1.1} & 3.8±1.8 \\
    \midrule
    \multirow{5}[12]{*}{} & No-graph & 32.2±0.2 & 10.31±2.1 & 31.3±0.5 & 18.0±0.5 & 14.7±0.6 \\
          & NDCN  & 8.2±0.5 & 6.1±2.4 & 6.6±0.7 & 4.4±0.5 & 9.1±0.8 \\
     II    & Adaptive-NDCN & 5.9±1.0 & 9.6±3.5 & 8.4±1.8 & 6.4±1.8 & 10.2±2.2 \\
      & GTS-NDCN & 6.5±1.2 & 9.0±2.1 & 10.4±1.8 & 9.4±3.2 & 8.4±3.2 \\
        \cmidrule{2-7}  & \textbf{SGODEv1} & 6.1±0.7 & \underline{5.6±2.5} & \underline{5.7±1.2} & \textbf{3.0±0.5} & \underline{7.7±2.7} \\
          & \textbf{SGODEv2} & \underline{5.1±0.6} & \textbf{4.2±1.3} & \textbf{5.2±1.3} & 4.8±1.0 & \textbf{6.2±2.1} \\
          & \textbf{SGODEv3} & \textbf{4.7±1.3} & 7.9±2.9 & 5.8±0.5 & \underline{3.1±0.4} & 9.4±3.7 \\
    \midrule
    \multirow{5}[12]{*}{} & No-graph & 26.9±0.1 & 11.5±0.2 & 25.2±1.1 & 15.7±0.1 & 18.2±0.3 \\
          & NDCN  & 5.9±1.6 & 3.3±0.3 & 3.1±0.2 & 4.0±0.4 & \textbf{1.9±0.4} \\
     III     & Adaptive-NDCN & 4.5±1.0 & 1.6±0.8 & 3.1±0.8 & 2.6±0.9 & 2.8±0.8 \\
     & GTS-NDCN & 4.9±1.2 & 2.7±0.6 & 6.1±0.9 & 6.2±1.3 & 3.4±0.4 \\
        \cmidrule{2-7}  & \textbf{SGODEv1} & 3.3±0.8 & \underline{2.3±0.8} & \textbf{2.0±0.2} & 2.8±0.6 & 2.6±0.9 \\
         & \textbf{SGODEv2} & \textbf{2.4±0.5} & \textbf{1.5±0.4} & 2.8±0.7 & \textbf{2.3±0.7} & 4.6±1.4 \\
         & \textbf{SGODEv3} & \underline{2.5±0.6} & 2.5±1.2 & \underline{2.0±0.3} & \underline{2.4±0.7} & \underline{2.7±1.1} \\
    \bottomrule
    \end{tabular}}
    \caption{MAPE of interpolation of continuous-time network dynamics. I: heat diffusion dynamics; II: mutualistic interaction dynamics; III: gene regulatory dynamics; Com.: Community. Those in black font indicate the best performance. The underline corresponds to the second-ranked value.}
     \label{tab:2}
     \vskip -0.2in
\end{table}

\begin{table}[htbp]
  \centering
  \resizebox{\linewidth}{!}{
    \begin{tabular}{c|ccccccccc}
    \toprule
    \multirow{2}[2]{*}{METR-LA} & \multicolumn{3}{c}{15min} & \multicolumn{3}{c}{30min} & \multicolumn{3}{c}{60min} \\
          &  MAE  &  RMSE & MAPE  &  MAE  &  RMSE & MAPE  &  MAE  &  RMSE & MAPE \\
    \midrule
    HA    & 4.16  & 7.80  & 13.0\% & 4.16  & 7.80  & 13.0\% & 4.16  & 7.80  & 13.0\% \\
    ARIMA & 3.99  & 8.21  & 9.6\% & 5.15  & 10.45 & 12.7\% & 6.90  & 13.23 & 17.4\% \\
    VAR   & 4.42  & 7.89  & 10.2\% & 5.41  & 9.13  & 12.7\% & 6.52  & 10.11 & 15.8\% \\
    SVR   & 3.99 & 8.45  & 9.3\% & 5.05  & 10.87 & 12.1\% & 6.72  & 13.76 & 16.7\% \\
    FNN   & 3.99 & 7.94  & 9.9\% & 4.23  & 8.17  & 12.9\% & 4.49  & 8.69  & 14.0\% \\
    LSTM  & 3.44  & 6.30  & 9.6\% & 3.77  & 7.23  & 10.9\% & 4.37  & 8.69  & 13.2\% \\
    DCRNN & 2.77  & 5.38  & 7.3\% & 3.15  & 6.45  & 8.8\% & 3.60  & 7.59  & 10.5\% \\
    GraphWaveNet & 2.69  & 5.15  & 6.9\% & 3.07  & 6.22  & 8.4\% & 3.53  & 7.37  & 10.0\% \\
    LDS   & 2.75  & 5.35  & 7.1\% & 3.14  & 6.45  & 8.6\% & 3.63  & 7.67  & 10.3\% \\
    MTGNN & 2.69  & 5.18  & 6.9\% & 3.05  & 6.17  & 8.2\% & 3.49  & 7.23  & 9.9\% \\
    GTSv  & 2.74  & 5.09  & 7.3\% & 3.11  & 6.02  & 8.7\% & 3.53  & 6.84  & 10.3\% \\
    GTS   & 2.64  & 4.95  & 6.8\% & 3.01  & 5.85  & 8.2\% & 3.41  & 6.74  & 9.9\% \\
    STEP  & 2.61  & 4.98  & 6.6\% & \underline{2.96}  & 5.97  & 8.0\% & \underline{3.37}  & 6.99  & 9.6\% \\
    MegaCRN & \underline{2.60}   & \underline{4.81}  & \underline{6.4\%} & 3.00  & \underline{5.74}  & \underline{7.8\%} & 3.41  & \underline{6.74}  & \underline{9.4\%} \\
    \midrule
    SGODE-RNN & \textbf{2.54} & \textbf{4.74} & \textbf{6.3\%} & \textbf{2.93} & \textbf{5.66} & \textbf{7.7\%} & \textbf{3.33} & \textbf{6.58} & \textbf{9.3\%} \\
    \midrule
    \multicolumn{1}{c}{\multirow{2}[2]{*}{PEMS-BAY}} & \multicolumn{3}{c}{15min} & \multicolumn{3}{c}{30min} & \multicolumn{2}{c}{60min} &  \\
          &  MAE  &  RMSE & MAPE  &  MAE  &  RMSE & MAPE  &  MAE  &  RMSE & MAPE \\
    \midrule
    HA    & 2.88  & 5.59  & 6.8\% & 2.88  & 5.59  & 6.8\% & 2.88  & 5.59  & 6.8\% \\
    ARIMA & 1.62  & 3.30  & 3.5\% & 2.33  & 4.76  & 5.4\% & 3.38  & 6.50  & 8.3\% \\
    VAR   & 1.74  & 3.16  & 3.6\% & 2.32  & 4.25  & 5.0\% & 2.93  & 5.44  & 6.5\% \\
    SVR   & 1.85  & 3.59  & 3.8\% & 2.48  & 5.18  & 5.5\% & 3.28  & 7.08  & 8.0\% \\
    FNN   & 2.20  & 4.42  & 5.2\% & 2.30  & 4.63  & 5.4\% & 2.46  & 4.98  & 5.9\% \\
    LSTM  & 2.05  & 4.19  & 4.8\% & 2.20  & 4.55  & 5.2\% & 2.37  & 4.96  & 5.7\% \\
    DCRNN & 1.39  & 2.95  & 2.9\% & 1.74  & 3.97  & 3.9\% & 2.07  & 4.74  & 4.9\% \\
    GraphWaveNet & 1.30  & 2.74  & \underline{2.7\%} & \underline{1.63}  & 3.70  & 3.7\% & 1.95  & 4.52  & 4.6\% \\
    LDS   & 1. 33 & 2.81  & 2.8\% & 1.67  & 3.80  & 3.8\% & 1.99 & 4.59  & 4.8\% \\
    MTGNN & \underline{1.32}  & 2.79  & 2.8\% & 1.65  & 3.74  & 3.7\% & 1.94  & 4.49  & 4.5\% \\
    GTSv  & 1.35  & 2.64  & 2.9\% & 1. 69 & 3.45  & 3.9\% & 1.99  & 4.05  & 4.7\% \\
    GTS   & \underline{1.32}  & \underline{2.62}  & 2.8\% & 1.64  & 3.41  & \textbf{3.6\%} & \underline{1.91}  & 3.97  & \textbf{4.4\%} \\
    STEP  & 1.36  & 2.73  & 2.8\% & 1.67  & 3.58  & 3.6\% & 1.99  & 4.20  & 4.6\% \\
    MegaCRN & 1.33  & 2.59  & 2.8\% & 1.65  & \underline{3.40}  & \underline{3.6\%} & 1.92  & \underline{4.04}  & \underline{4.5\%} \\
    \midrule
    SGODE-RNN & \textbf{1.29} & \textbf{2.55} & \textbf{2.7\%} & \textbf{1.61} & \textbf{3.35} & \textbf{3.6\%} & \textbf{1.90} & \textbf{3.96} & \textbf{4.4\%} \\
    \bottomrule
    \end{tabular}}
    \vskip -0.1in
      \caption{Forecasting error on METR-LA and PEMS-BAY.}
  \label{tab:metr}
  \vskip -0.15in
\end{table}

\subsection{Results}
\paragraph{Results on Three Dynamics} The purpose of synthetic data experiment is to test the performance of different graph strategies (GTS: graph distribution inference, Adaptive: adaptive graph without negative information, NDCN: real graph) based on a unified base model. 
We conducted experiments involving both internal and external interpolation within continuous-time dynamics. The outcomes of our internal interpolation experiment are detailed in Table \ref{tab:2}, while those from the external interpolation experiment are available in the Appendix. SGODE adeptly captures the evolution of various nodes across time and exhibits strong performance across diverse graph dynamics. Notably, methods employing learnable graphs are at least on par with NDCN when using the actual graph and, importantly, outperform approaches that lack a graph structure.

In the interpolation experiment, our method demonstrates superior accuracy compared to other techniques, particularly the graph inference methods Adaptive-NDCN and GTS-NDCN. Conversely, the NDCN variant lacking a graph exhibits a significant loss value, indicating its incapability to learn dynamic models and underscoring the importance of aggregating neighborhood information. In extrapolation experiments, SGODE exhibits substantial superiority over No-graph, Adaptive, GTS, and NDCN in heat diffusion dynamics. Our method also consistently achieves top-tier outcomes in most extrapolation cases. The utilization of node embedding vectors to approximate the coefficient matrix, as opposed to directly learning the entire matrix, maintains accuracy without compromise. The marginal underperformance of our approach relative to NDCN in certain network dynamics may arise from the fact that these dynamics are generated from ground-truth graphs. By incorporating SGODE, we enhance the precision of the ODE-GNN model in predictive dynamics, especially in the context of interpolation prediction.

\paragraph{Results on Traffic Datasets}

\textit{Forecasting quality.} For prediction accuracy, we compare SGODE-RNN with the previously mentioned methods. Following the conventions of DCRNN \cite{li2018diffusion}, GTS \cite{shang2021discrete}, and STG-NCDE \cite{choi2022graph}, we present results for 15min, 30min, and 60min on METR-LA and PEMS-BAY datasets in Table \ref{tab:metr}, while average outcomes on PeMSD4 and PeMSD8 are reported in Table \ref{tab:pems}.

We have observed the following phenomena: 1) In Fig. \ref{fig:visualization}, the results demonstrates that our proposed approach is capable of capturing signed information. 2) On the METR-LA and PEMS-BAY datasets, SGODE significantly outperforms DCRNN, and it also surpasses state-of-the-art traffic flow prediction methods. This showcases the effectiveness of our proposed strategy in leveraging signed information and the positive benefits of embedding such information.

Since our approach relies on ODEs, we compare our experimental results with STNODE and STG-NCDE, as shown in Table \ref{tab:pems}. The results of our experiments convincingly showcase the efficacy of our method, while our variants further unlock the potential of the STG-NCDE model. Moreover, our proposed SGODE-DCRNN demonstrates remarkable competitiveness. Notably, it is observed that the MAPE and MSE metrics of GTS and SGODE-DCRNN do not align with the MAE metric on the PeMSD4 dataset, possibly due to the presence of some minima close to zero in this dataset. 

\begin{figure}[htb]
\vskip -0.1in
\centering
\subfloat[$\boldsymbol{K}_{top5}$] {\includegraphics[width=0.22\textwidth]{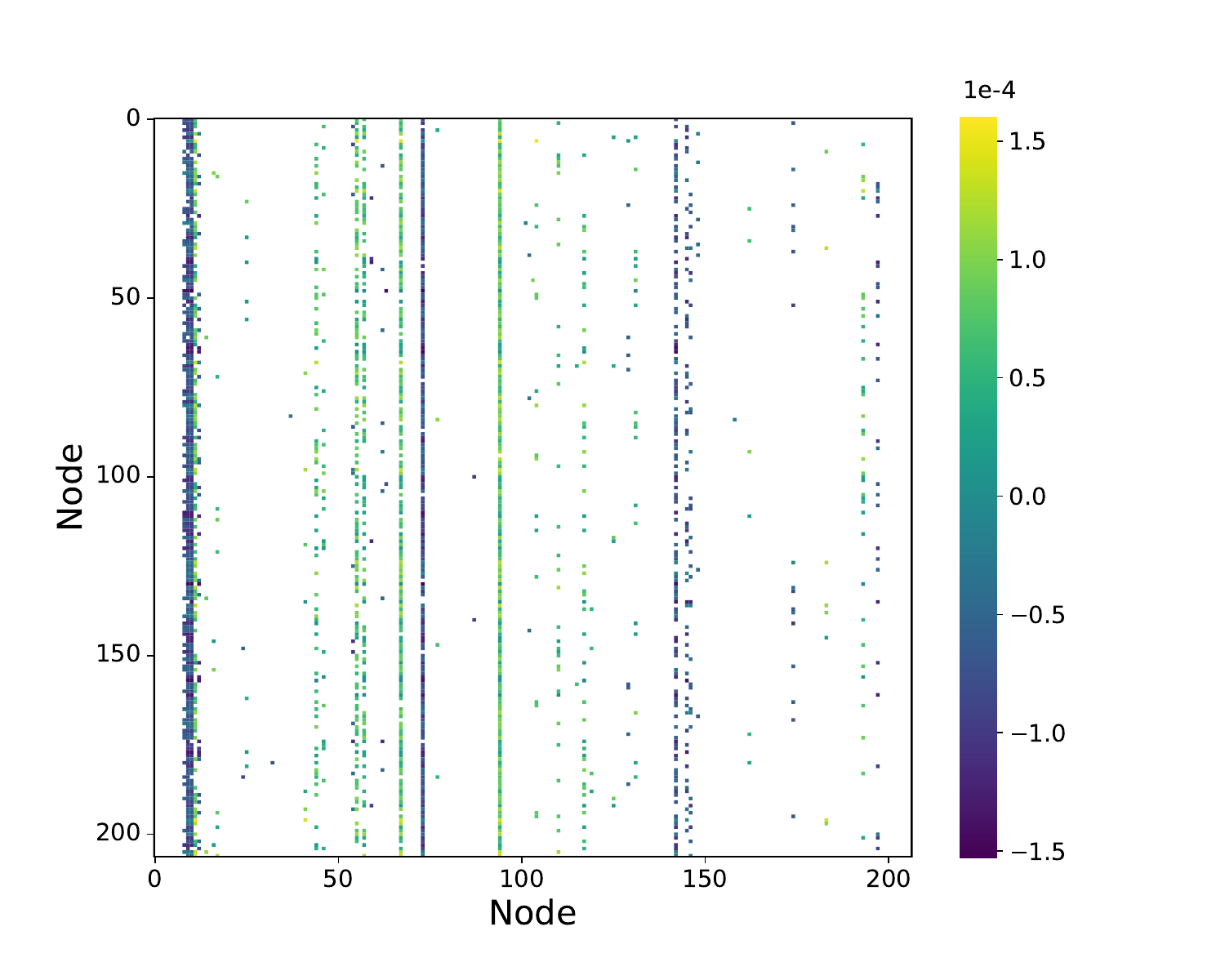}}\label{figtop5}
\subfloat[Important Nodes] {\includegraphics[width=0.22\textwidth]{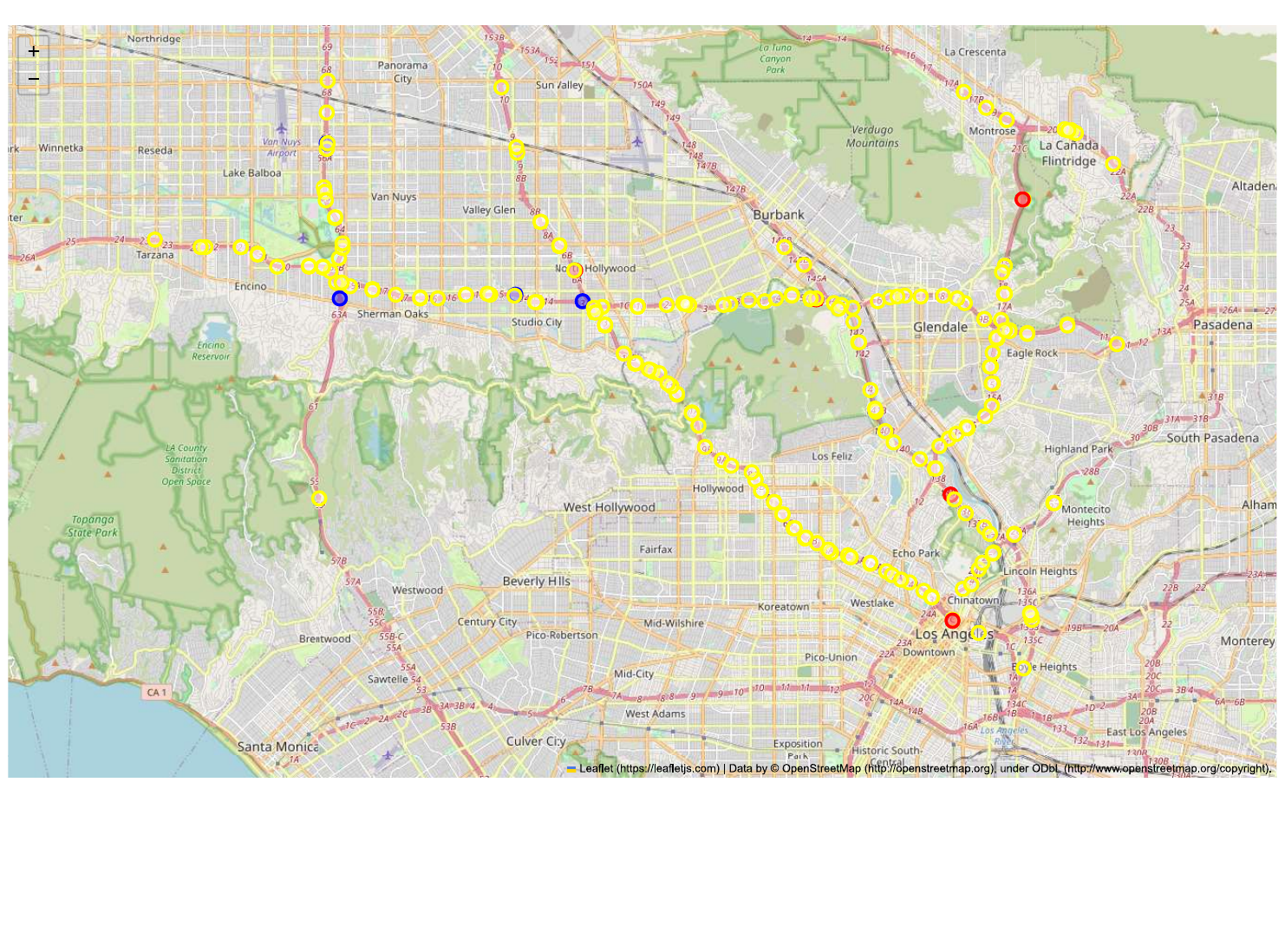}\label{figgraphtop5}}
\vskip -0.1in
\caption{Visualization of $\boldsymbol{K}_{top5}$ and important nodes of SGODE-RNN in METR-LA. 
The term "crucial nodes" refers to nodes in $\boldsymbol{K}_{top5}$ with degrees greater than $1/2N$ after applying the sign operation.
(a) $\boldsymbol{K}_{top5}$ refers to the top five ranked sum of absolute weights in the row vector corresponding to each node in matrix $\boldsymbol{K}$. 
(b) The important nodes are distinguished by red (positive) and blue (negative) colors. Additional visualizations can be found in the Appendix.
}
\label{fig:visualization}
\vskip -0.2in
\end{figure}

\begin{table}[htbp]
\vskip -0.1in
  \centering
  \resizebox{\linewidth}{!}{
    \begin{tabular}{ccccccc}
    \toprule
    \multirow{2}[4]{*}{Model} & \multicolumn{3}{c}{PeMSD4} & \multicolumn{3}{c}{PeMSD8} \\
\cmidrule{2-7}          & MAE   & RMSE  & MAPE  & MAE   & RMSE  & MAPE \\
    \midrule
    HA    & 38.03 & 59.24 & 27.88\% & 34.86 & 59.24 & 27.88\% \\
    ARIMA & 33.73 & 48.80 & 24.18\% & 31.09 & 44.32 & 22.73\% \\
    VAR   & 24.54 & 38.61 & 17.24\% & 19.19 & 29.81 & 13.10\% \\
    DCRNN & 21.22 & 33.44 & 14.17\% & 16.82 & 26.36 & 10.92\% \\
    GraphWaveNet & 24.89 & 39.66 & 17.29\% & 18.28 & 30.05 & 12.15\% \\
    STGODE & 20.84 & 32.82 & 13.77\% & 16.81 & 25.97 & 10.62\% \\
    STG-NCDE & 19.21 & \underline{31.09} & \underline{12.76}\% & 15.45 & 24.81 & 9.92\% \\
    GTS   & 19.36 & 32.99 & 13.54\% & \underline{14.82} & \textbf{23.80}  & \underline{9.52}\% \\
    MegaCRN & 18.92 & 31.90 & 12.89\% & 14.91 & 23.97 & 9.60\%\\
    \midrule
    SGODE-NCDE & \underline{19.06} & \textbf{30.96} & \textbf{12.65\%} & 15.34 & 24.44 & 9.92\% \\
    SGODE-RNN & \textbf{18.81} & 31.57 & 12.87\% & \textbf{14.55} & \underline{23.85} & \textbf{9.30\%} \\
    \bottomrule
    \end{tabular}}
    \vskip -0.1in
    \caption{Forecasting error on PeMSD4 and PeMSD8.}
    \vskip -0.2in
  \label{tab:pems}
\end{table}

In practice, data recording and storage errors can lead to the unavailability of specific data \cite{choi2022graph}. We compare SGODE-NCDE and STG-NCDE on PeMSD4 and PeMSD8 datasets with 10\%, 30\%, and 50\% data corruption, and the results are presented in Table \ref{tab:miss-04}.
Our experimental setup is based on STG-NCDE. The results indicate that our approach exhibits greater robustness to irregular datasets compared to STG-NCDE. This underscores the accuracy of our method in fitting continuous dynamics.

\begin{table}[htbp]
\vskip -0.1in
  \centering
  \resizebox{\linewidth}{!}{
    \begin{tabular}{cccccccccc}
    \toprule
    \multirow{2}[2]{*}{PeMSD4} & \multicolumn{3}{c}{Missing rate(10\%)} & \multicolumn{3}{c}{Missing rate(30\%)} & \multicolumn{3}{c}{Missing rate(50\%)} \\
   \cmidrule{2-10}   
          & MAE   & RMSE  & MAPE  & MAE   & RMSE  & MAPE  & MAE   & RMSE  & MAPE \\
    \midrule
    STG-NCDE & 19.61 & 31.55 & 13.02\% & \textbf{19.37} & \textbf{31.27} & 13.12\% & 20.35 & 32.41 & 13.50\% \\
    SGODE-NCDE  & \textbf{19.15} & \textbf{31.05} & \textbf{12.96\%} & 19.41 & 31.29 & \textbf{12.86\%} & \textbf{19.70} & \textbf{31.86} & \textbf{13.07\%} \\
    \midrule
    \multirow{2}[4]{*}{PeMSD8} & \multicolumn{3}{c}{Missing rate(10\%)} & \multicolumn{3}{c}{Missing rate(30\%)} & \multicolumn{3}{c}{Missing rate(50\%)} \\
\cmidrule{2-10}          & MAE   & RMSE  & MAPE  & MAE   & RMSE  & MAPE  & MAE   & RMSE  & MAPE \\
    \midrule
    STG-NCDE & 17.21 & 26.96 & 10.68\% & 17.42 & 27.48 & 11.03\% & 17.21 & 26.93 & 11.02\% \\
    SGODE-NCDE  & \textbf{15.67} & \textbf{24.95} & \textbf{10.05\%} & \textbf{16.16} & \textbf{25.15} & \textbf{10.86\%} & \textbf{15.88} & \textbf{25.11} & \textbf{10.33\%} \\
    \bottomrule
    \end{tabular}}
    \vskip -0.1in
    \caption{Forecasting error on irregular PeMSD4 and PeMSD8.}
  \label{tab:miss-04}
\end{table}
\vskip -0.15in
\begin{figure}[htb]
\vskip -0.2in
\centering
\subfloat[MAE on PeMSD4] {\includegraphics[width=0.24\textwidth]{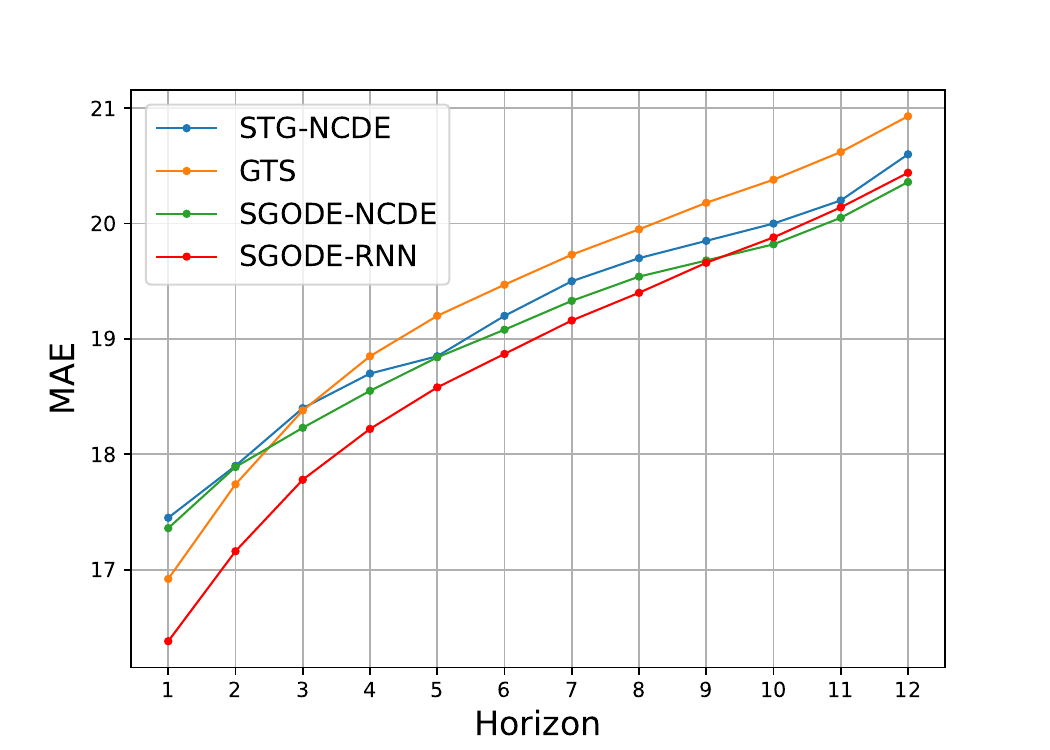}\label{fighorizon(a)}}
\subfloat[MAE on PeMSD8] {\includegraphics[width=0.24\textwidth]{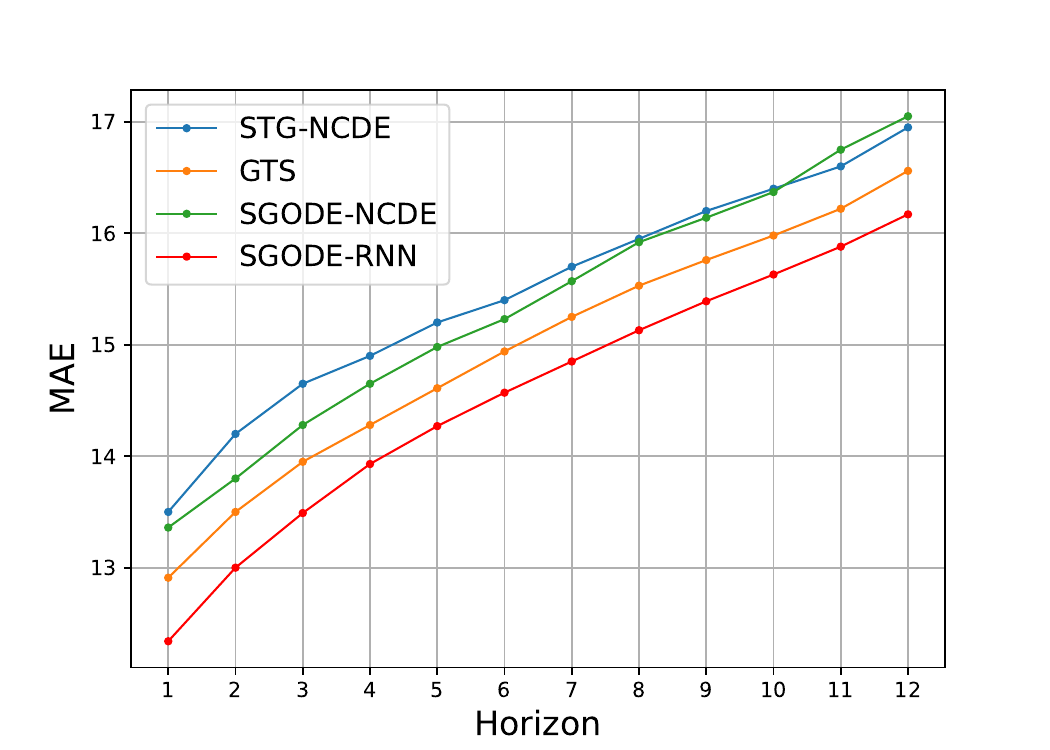}\label{fighorizon(b)}}
\vskip -0.1in
\caption{Prediction error at each horizon. More results in
other datasets are in Appendix.}
\label{fig:error}
\vskip -0.1in
\end{figure}

\textit{Error for each horizon.}
We present the error analysis of each horizon prediction in Fig.\ref{fig:error}, where we predict a total of 12 horizons. It is evident that the level of error is highly correlated with forecast time. Across all horizons, SGODE-RNN outperforms both baseline models GTS and STG-NCDE in terms of lower error rates. While our proposed methods SGODE-NCDE and SGODE-RNN exhibit superior performance in the initial few horizons, their advantage over STG-NCDE and GTS gradually diminishes as forecasting progresses.

\subsection{Ablation Study}\label{ablation}
We verify the validity of our proposed two key components, the designed $\boldsymbol{K}$ and $\boldsymbol{B}$. We denote the SGODE-RNN with negative links removed and without its trend $\boldsymbol{B}$ as \textit{Positive1} and \textit{without}-$\boldsymbol{B}$, respectively. \textit{Only FF} performs feature extraction solely on the final ODE result, without extracting features from the initial state and intermediate states. \textit{Positive2} refers to $\boldsymbol{K}=\boldsymbol{K}_{pos_1}+\boldsymbol{K}_{pos_2}$. We show the results on four datasets in Table \ref{tab:Ablation experiment}.

The experimental results demonstrate a significant performance decrease of SGODE when our strategies are not considered, except for the PEMS-BAY dataset, which exhibits a relatively minor improvement in performance related to signed relationships. In comparison to \textit{Positive2}, we also eliminate the possibility of performance improvement due to the introduction of additional parameters.

\begin{table}[htbp]
\vskip -0.1in
  \centering
  \resizebox{\linewidth}{!}{
    \begin{tabular}{ccccccccccccc}
    \toprule
    \multirow{2}[3]{*}{Model} & \multicolumn{3}{c}{PeMSD4} & \multicolumn{3}{c}{PeMSD8} & \multicolumn{3}{c}{METR-LA} & \multicolumn{3}{c}{PEMS-BAY} \\
\cmidrule{2-13}          & MAE   & RMSE  & MAPE  & MAE   & RMSE  & MAPE  & MAE   & RMSE  & MAPE  & MAE   & RMSE  & MAPE \\
\midrule
    \textit{Only FF} & 19.32 & 33.03 & 13.05\% & 15.07 & 24.41 & 10.05\% & 2.92  & 5.64  & 7.74\% & 1.57  & 3.19  & 3.53\% \\
    
    \textit{Without}-$\boldsymbol{B}$ & 19.01 & 31.31 & 14.13\% & 15.18 & 24.15 & 9.97\% & 2.95  & 5.71  & 8.05\% & 1.58  & 3.23  & 3.51\% \\
    \textit{Positivev1} & 19.24 & 31.91 & 13.16\% & 14.81 & 23.95 & 9.57\% & 2.89  & 5.60  & 7.81\% & 1.57  & 3.19  & 3.54\% \\
    \textit{Positivev2} & 19.56 & 32.69 & 13.79\% & 15.76 & 24.84 & 11.49\% & 2.93  & 5.69  & 8.13\% & 1.57  & 3.22  & 3.58\% \\  
    SGODE & \textbf{18.81} & \textbf{31.57} & \textbf{12.87\%} & \textbf{14.55} & \textbf{23.85} & \textbf{9.30\%} & \textbf{2.88} & \textbf{5.52} & \textbf{7.61\%} & \textbf{1.55} & \textbf{3.15} & \textbf{3.50\%} \\
    \bottomrule
    \end{tabular}}
    \vskip -0.1in
    \caption{Ablation study of SGODE-RNN.}
  \label{tab:Ablation experiment}
  \vskip -0.2in
\end{table}

\subsection{Parameters Analysis}

We illustrate the impact of the diffusion steps $m$ and the second dimension $d$ of the node embedding matrix $\boldsymbol{E}_{*}$ in Fig. \ref{fig:para_METR_LA}. Notably, higher diffusion steps (greater than 1) yield improved outcomes, suggesting the practicality of extracting intermediate states. The model's accuracy initially rises with increasing node embedding dimension, followed by a gradual decline as the dimension further increases. As a result, we advise opting for a manageable dimension when making your selection.

\begin{figure}[htb]
\vskip -0.2in
\centering
\subfloat[Fix dim = 10 and vary $m$.] {\includegraphics[width=0.25\textwidth]{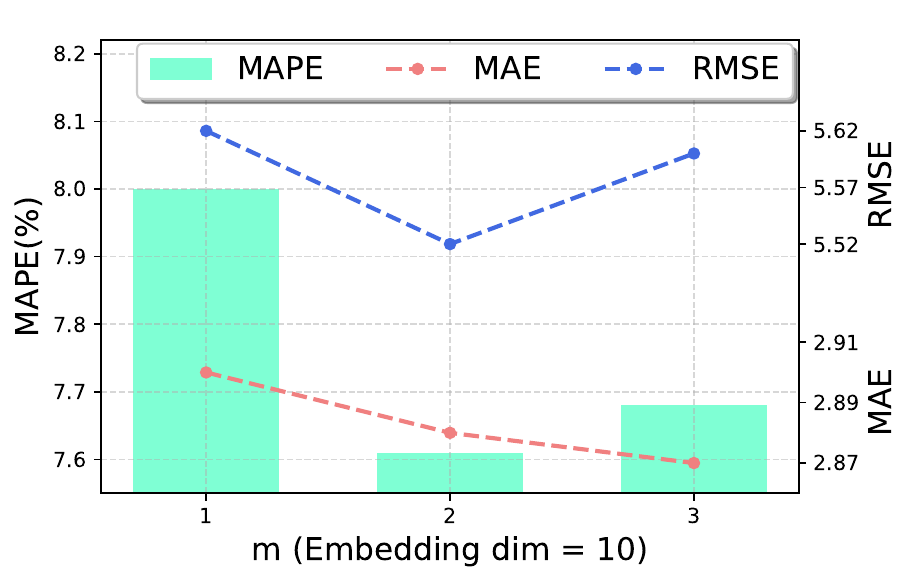}\label{figpa(a)}}
\subfloat[Fix $m$ = 2 and vary dim.] {\includegraphics[width=0.25\textwidth]{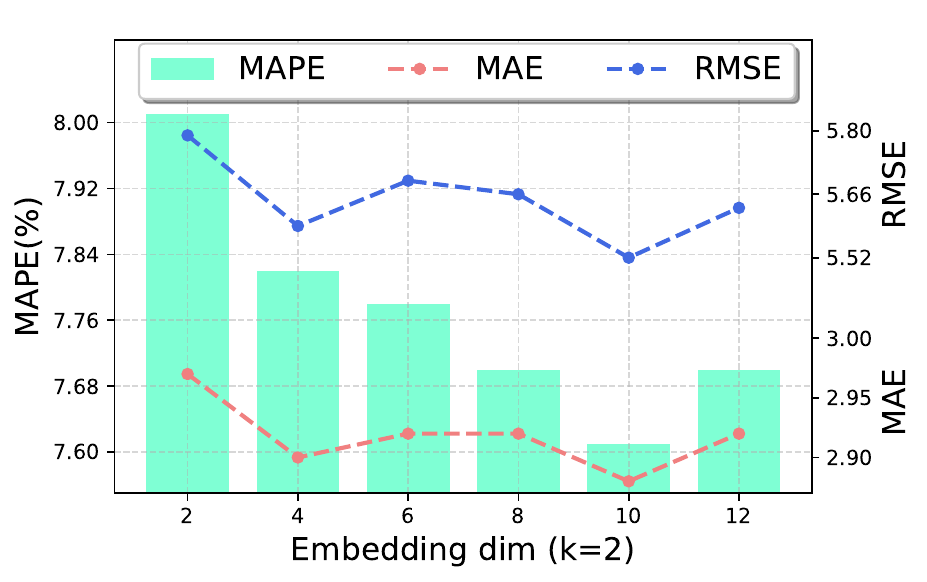}\label{figpa(b)}}
\vskip -0.1in
\caption{Sensitivity analysis of $m$ and $d$ on METR-LA. More results in other datasets are shown in Appendix.}
\label{fig:para_METR_LA}
\vskip -0.1in
\end{figure}

\begin{figure}[htb]
\vskip -0.1in
\centering
\subfloat[Parametric efficiency] {\includegraphics[width=0.25\textwidth]{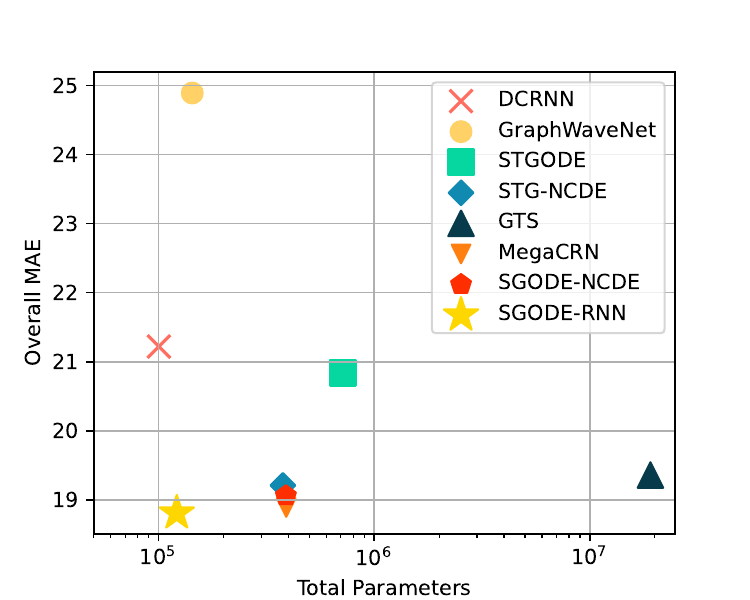}\label{figeff(a)}}
\subfloat[Runtime Efficiency] {\includegraphics[width=0.25\textwidth]{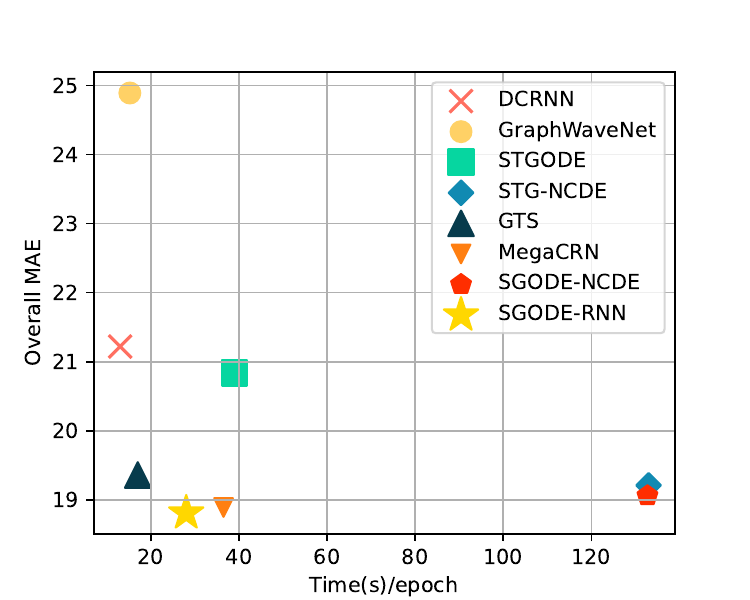}\label{figeff(b)}}
\vskip -0.1in
\caption{Efficiency Evaluation on PEMS04.}
\label{fig:parameters and time}
\vskip -0.1in
\end{figure}
\subsection{Efficiency Study}
We assess the effectiveness of our approaches through comparisons with state-of-the-art methods. Fig. \ref{fig:parameters and time} showcases the efficiency of our methods in terms of parameters and runtime, compared to the state-of-the-art alternatives.
Our methods (SGODE-RNN) achieve the second lowest parameter count, yet attain the minimum overall MAE. Additionally, our approach demonstrates reasonable runtime performance. Notably, methodologies incorporating ODE solvers result in extended training durations, while the supplementary runtime introduced by SGODE (SGODE-NCDE) remains minimal.

\section{Related Work}\label{related work}
Lately, a noteworthy trend has emerged that involves the fusion of ODEs with graph neural networks to acquire insights into continuous-time dynamics. 
Furthermore, a concerted effort has been dedicated to incorporating graph structures into dynamic systems. 
Noteworthy endeavors have been made in this domain to extend NODE \cite{NEURIPS2018_NODE} to harness the rich graph structure \cite{Hwang2021Climate,poli2019graph}.
GDE \cite{poli2019graph} extends the reach of GNNs into the continuous field by leveraging ODEs to learn input-output relationships. CGNN \cite{xhonneux2020continuous} introduces a continuous message-passing layer, defining derivatives as combined representations of both the current and initial nodes. LG-ODE \cite{huang2020learning} employs neighborhood information to gather dynamic contextual cues, addressing scenarios where node states may not be observable over time. NDCN \cite{Zang2020Neural} ingeniously combines ODEs and GNNs for modeling continuous-time dynamics. STGODE \cite{Fang2021Spatial} captures spatio-temporal dynamics using tensor-based ODEs. Moreover, a distinct strand of research endeavors to extend ODEs to the realm of dynamic graphs. CG-ODE \cite{huang2021coupled} integrates coupled ODEs to model dynamics based on edges and nodes, respectively. Jin \emph{et al.} \cite{jin2022neural} introduce explicit temporal dependencies through ODEs, showcasing their efficacy in graph-based modeling tasks and underlining the significance of graph representation. Nevertheless, the presumption of a known exact graph structure in advance is often challenging to maintain. 
STG-NCDE \cite{choi2022graph} introduces a dynamic approach with two NCDEs, adeptly handling spatial and temporal information using an adaptive normalized adjacency matrix. In a similar vein, MTGODE \cite{jin2022neural} abstracted input sequences into dynamic graphs, where node features evolve over time alongside an unspecified graph structure.

\section{Conclusions}
We introduce a simple yet effective framework and extend its application to more intricate scenarios. Our investigation substantiates the inherent capability of the proposed methodology to effectively capture and leverage signed information. Furthermore, we underscore the method's adaptability, showcasing its effortless integration into a diverse range of cutting-edge dynamic modeling methodologies. Through comprehensive experimentation encompassing both synthetic and real-world datasets, we unveil the untapped potential of incorporating signed relations. This integration results in notable enhancements in performance, particularly in the context of short-term or interpolation prediction tasks. However, in cases where the graph size is substantial, challenges arise in accurately learning the signed graph, potentially leading to overfitting to local optima.

\section{Acknowledgments}


This work was supported in part by the National Natural Science Foundation of China under Grant 62206205, 62206207, in part by the Guangdong High-level Innovation Research Institution Project under Grant 2021B0909050008, and in part by the Guangzhou Key Research and Development Program under Grant 202206030003.
\bibliography{aaai24}

\newpage
\appendix 
\begin{figure*}[htbp]
\centering
\subfloat[] {\includegraphics[width=0.25\textwidth]{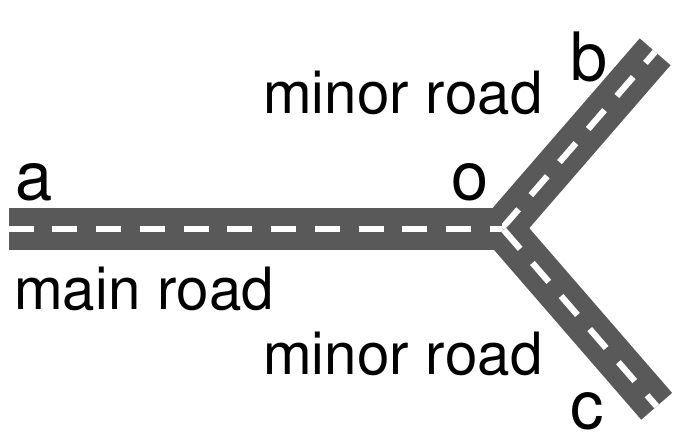}\label{fig1(a)}}
\subfloat[] {\includegraphics[width=0.25\textwidth]{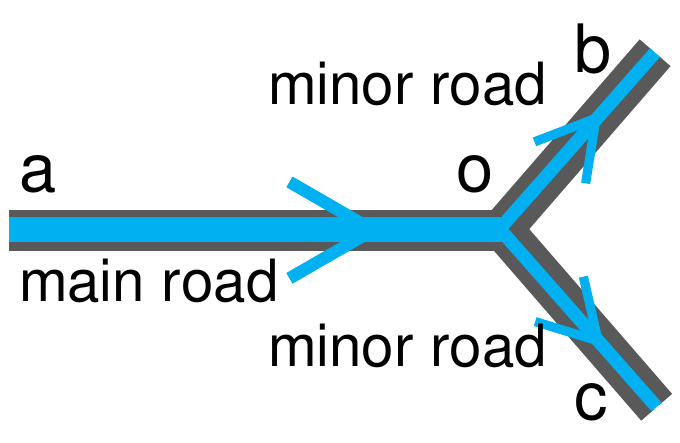}\label{fig1(b)}}
\subfloat[] {\includegraphics[width=0.25\textwidth]{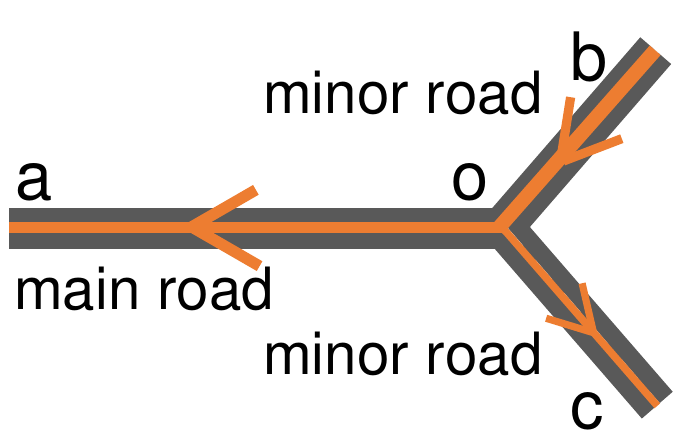}\label{fig1(c)}}
\subfloat[] {\includegraphics[width=0.25\textwidth]{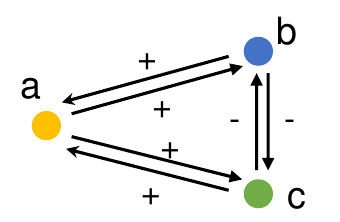}\label{fig1(d)}}
\caption{The relationships of traffic flow among three nodes at the three-fork intersection, including main road $a$, minor road $b$ and minor road $c$. We assume that the traffic flow on the main road $a$ is larger compared to the minor roads, represented by the blue line of $aob$ and $aoc$ in (b). And the traffic flow from the minor road to the main road, represented by the orange line of $boa$ in (c), is much larger than the traffic flow on the branch road to the top of the branch road, $boc$ in (c). 
Thus, as shown in (d), we can obtain the relationship, which is described by signed graph.}
\vskip -0.1in
\label{fig1}
\end{figure*}

\begin{figure*}[htbp]
\centering
\subfloat[] {\includegraphics[width=0.25\textwidth]{graphvisualization/METRLAtop5.pdf}\label{figgraph(a)}}
\subfloat[] {\includegraphics[width=0.25\textwidth]{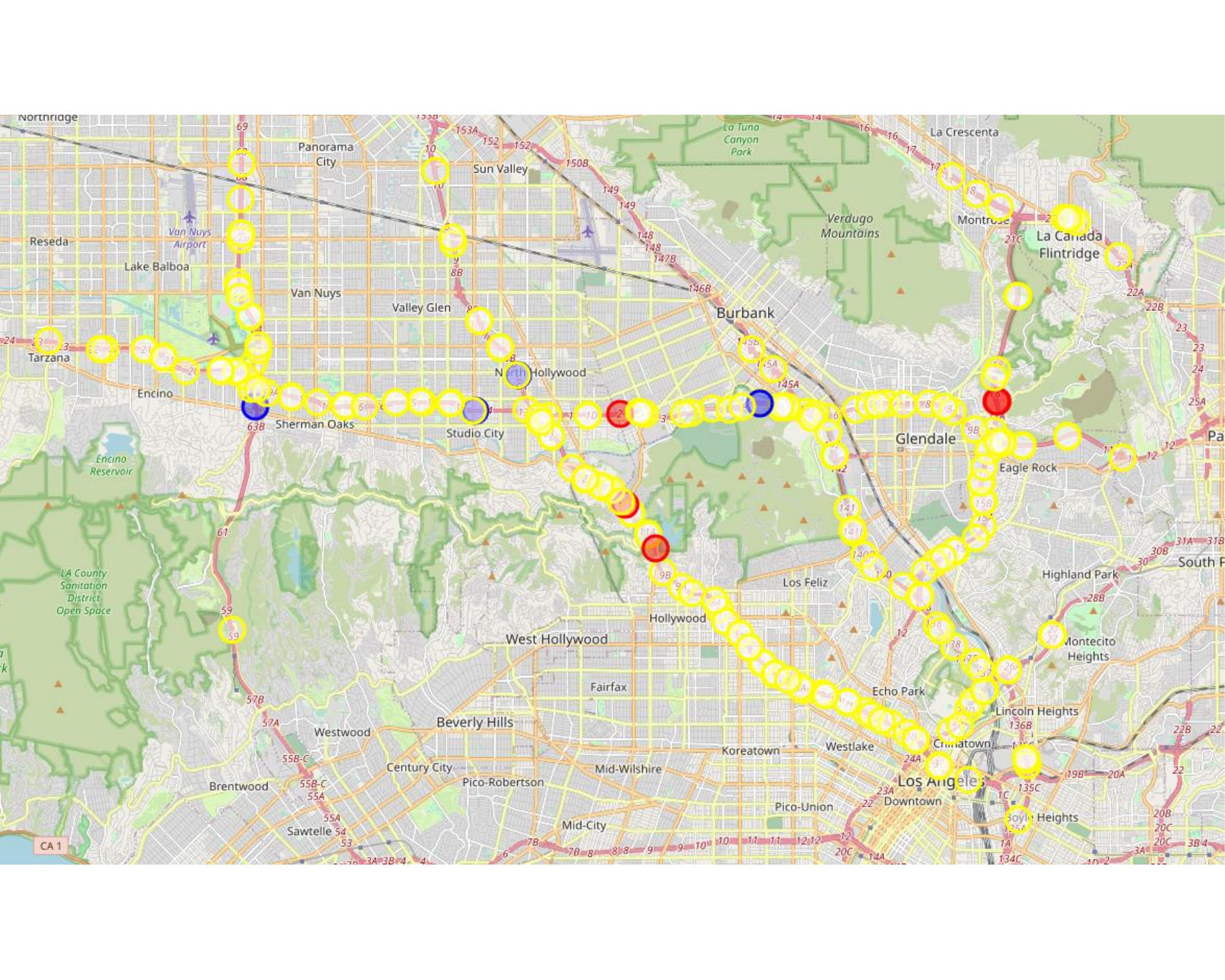}\label{figgraph(b)}}
\subfloat[] {\includegraphics[width=0.25\textwidth]{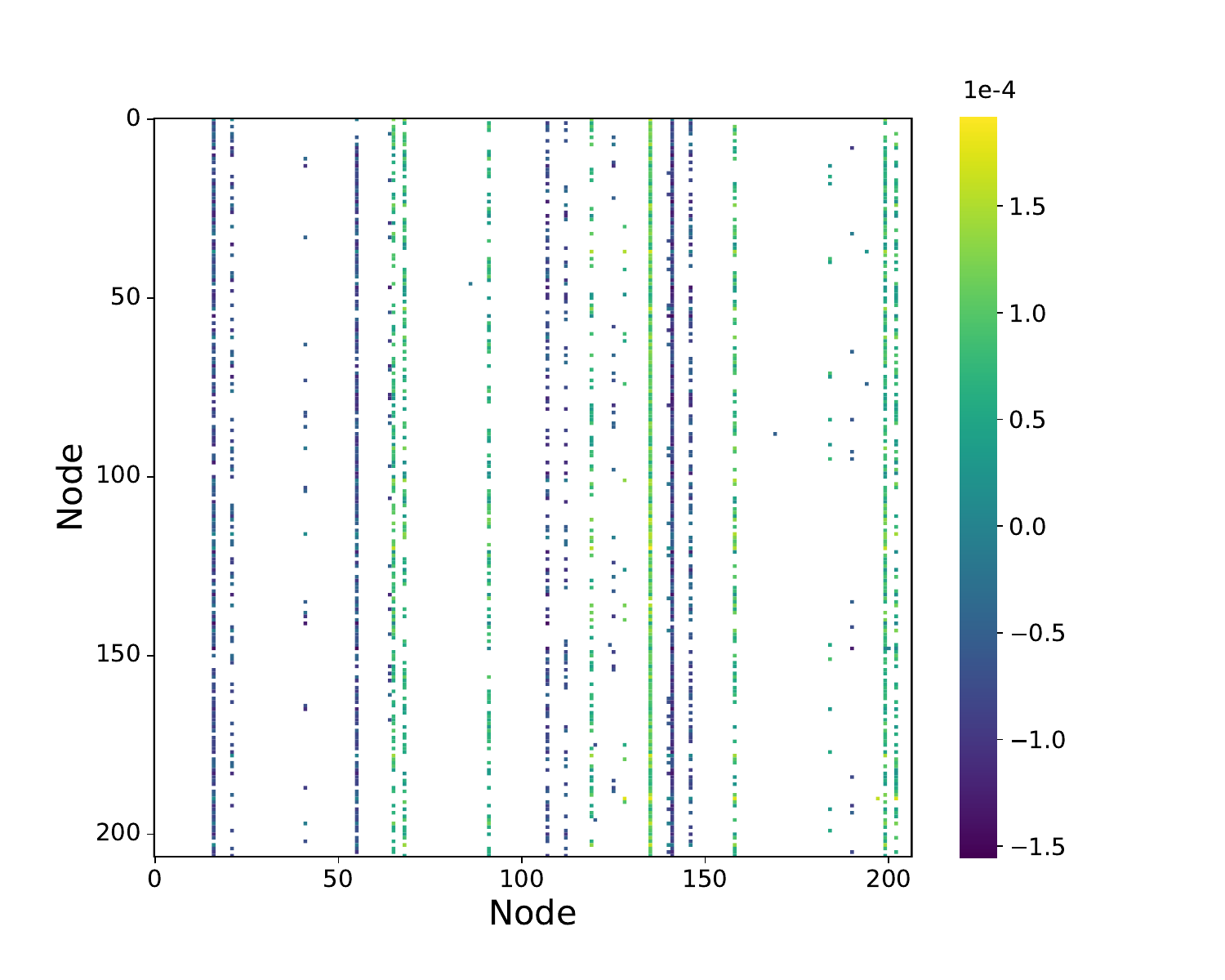}\label{figgraph(c)}}
\subfloat[] {\includegraphics[width=0.25\textwidth]{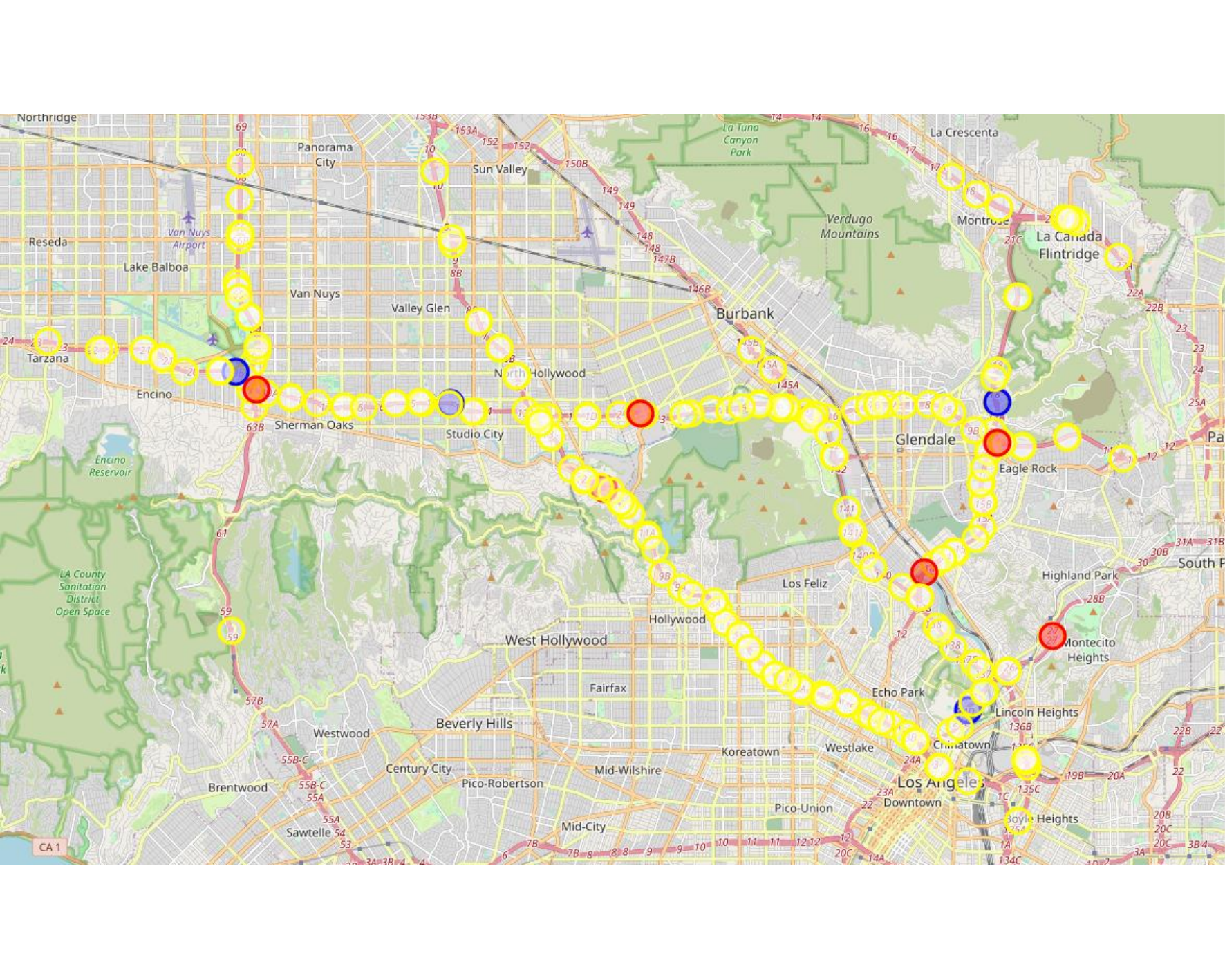}\label{figgraph(d)}}
\caption{Visualization of $\boldsymbol{K}{top5}$ and important nodes of SGODE-DCRNN in METR-LA. The term "important nodes" refers to nodes in $\boldsymbol{K}{top5}$ that, after undergoing the sign operation, exhibit degrees greater than $1/2N$. The sign operation entails mapping negative values to -1, positive values to 1, and leaving 0 values unchanged. Panels (a) and (c) illustrate $\boldsymbol{K}_{top5}$, representing the top 5 absolute weights in the row vector corresponding to each node in matrix $\boldsymbol{K}$. Panels (b) and (d) depict the important nodes, wherein negative important nodes are highlighted in blue and positive important nodes are highlighted in red.}
\vskip -0.1in
\label{figgraph}
\end{figure*}

\begin{figure*}[htbp]
\centering
\includegraphics[width=0.95\textwidth]{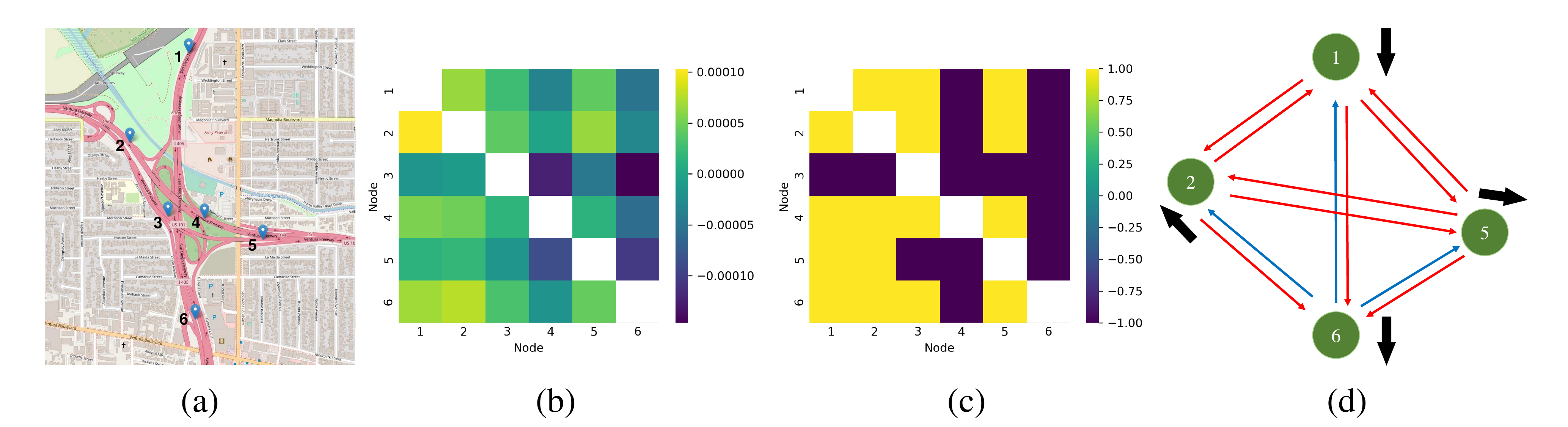}
\caption{Local visualization. We select a local intersection of the METR-LA dataset, and the six nodes in Fig. (a) in turn represent the 78, 132, 131, 159, 44, and 10 sensors in the dataset. Fig. (b) shows the learned coefficients (encoders), ignoring the weights on the diagonal. Fig. (c) is the result of signed graph(take -1 and 1 only). Fig. (d) is a visualization of Fig. (c), where the red arrow represents positive, the blue arrow represents negative, and the black bold arrow represents the direction of the intersection detected by the sensor.We denote by row $i$ and column $j$ the sign of the edge from node $j$ to node $i$ in Fig. (c).}
\vskip -0.1in
\label{figlocal}
\end{figure*}

The appendix is organized as follows. We first intoduce an example of a three-way intersection in traffic systems. Then we demonstrate the signed graph we learned and visualize. Next we report the hyperparameters of our experiments for easy reproduction. Furthermore we add the knowledge about the three synthetic dynamics and the way of data generation and report the results for irregularly sampled and regularly sampled extrapolation experiments. Lastly, we present the prediction error and sensitivity analysis for each horizon. For reproducibility, our code is available as a zip file in the supplementary material, and we will make our code publicly available in the future.


\section{Three-way intersection in traffic systems}

We take an example of a three-way intersection in traffic systems to show that there may also be negative relationships between traffic flows on different nodes as shown in Fig. \ref{fig1}.

It is evident that the connection of node $a$ to node $b$ and node $a$ to node $c$ are positive. For the relationship between node $b$ and $c$, we statically analyze the neighborhood information obtained by node $c$. First, assume that node $b$ does not exist, then the increment on node $a$ is the same as on node $c$. The effect on node $c$ after adding node $b$ is that a part of the traffic flow on node $a$ that should flow to node $c$ will flow to node $b$, while a part of the flow on node $b$ will flow to node $c$. However, the traffic flow $aob$ is much larger than the traffic flow $boc$, so the increment on node $b$ means the increment on node $c$ decreases, i.e., the link from node $b$ to node $c$ is negative. 

\section{Visualization and analysis}
In this section, we show the coefficient matrice $\boldsymbol{K}$ we learned in the encoder and decoder, and we choose a local region to analyze the positive and negative relationships we learned. The maps used in this article are obtained from the Python library folium\footnote{https://python-visualization.github.io/folium/index.html}.
\subsection{Global visualization}

Fig. \ref{figgraph} presents the visualization of $\boldsymbol{K}_{top5}$ and important nodes within the SGODE-DCRNN model applied to METR-LA. Specifically, we display the weight distribution of the first 5 and the last 5 elements within each row of the coefficient matrix. Additionally, we identify and mark important nodes on the graph, with significance defined based on the nodes' positive and negative degrees after applying the sign operation to matrix $K$. This insight is demonstrated in Fig. \ref{figgraph}. Our observations can be summarized as follows:

\begin{enumerate}
\item  \textbf{Attention to important Nodes}: The learned underlying signed graph highlights important nodes. Across our learned coefficient matrix, a majority of nodes exhibit pronounced attention toward select nodes, underscoring the substantial influence these important nodes wield within the transportation network.
\item  \textbf{Intersection Focus}: Notably, important nodes are often situated at intersections. Fig. \ref{figgraph(b)} and Fig. \ref{figgraph(c)} reveal a concentration of important nodes at intersections or along the edges of the map. Such nodes manifest as both positive and negative important nodes at intersections.
\item  \textbf{Encoder-Decoder Emphasis}: We observe a distinction in emphasis between the encoder and decoder. The encoder's coefficient matrix exhibits a bias toward the graph's top-left region, while the decoder's matrix leans toward the bottom-right portion. The latter aligns with the direction heading to Los Angeles, implying that during hidden vector extraction, the encoder emphasizes comprehensive information, favoring non-busy areas. In contrast, during result output, the decoder prioritizes important information, predominantly at intersections and bustling regions.
\end{enumerate}

\subsection{Local visualization }
We select the highway dataset as our chosen dataset, focusing on the traffic flow of complex interchanges with elevated roads. In this context, we analyze a local intersection to extract insights. Our analysis leads to two key observations. First, nodes with similar characteristics exhibit the same sign. For instance, Fig. \ref{figlocal}(a) illustrates that node 3 and node 5, situated on the same main road, share the same sign. This pattern holds true across the other nodes, as indicated in Fig. \ref{figlocal}(b) and (c) (columns 3 and 5). Second, the utilization of positive and negative relations provides a more accurate and reasonable description of road interconnections. For a lucid analysis, we specifically consider nodes 1, 2, 5, and 6, as illustrated in Fig. \ref{figlocal}(c). Nodes 2 and 5 correspond to sensors on two opposing directions of a road, one leading east to Los Angeles and the other west to Oxnard. Nodes 1 and 6 represent sensors on the southbound side of the road, stretching from Santa Clarita to Inglewood. This configuration suggests that nodes 2 and 5 are positioned on major traffic routes, while node 6 holds a relatively lesser significance. Analyzing the learned weights, we observe that the relationship between nodes 1, 2, and 5 remains consistent, while these nodes exhibit both positive and negative associations with node 6. This reinforces the adaptability of our method in comprehending intricate road relationships.

\section{Model hyperparameters}\label{hyper}
For the sake of reproducibility, we present the hyperparameters utilized in our models on datasets. All experiments are conducted in the following hardware and software environments: UBUNTU 20.04 LTS, PYTHON 3.9.7, NUMPY 1.21.2, SCIPY 1.7.3, PYTORCH 1.11.1, CUDA 11.3, NVIDIA Driver 470.103, i7 CPU, and NVIDIA GeForce RTX 3090. In all cases, we run ten times using different random seeds.
 
\begin{table*}[htbp]
  \centering
  \small
    \caption{Evolutionary time of dynamics}
      \begin{tabular}{rlllll}
    \toprule
    & Grid  & Random & Power Law & Small World & Community \\
    \midrule
 Heat Diffusion & 5 & 0.08 & 0.6 & 2 & 0.4 \\
    \midrule
Mutualistic Interaction  & 5 & 2 & 4 & 5 & 4 \\
    \midrule
       Gene Regulation   &5 & 4 & 1.5 & 4.5 & 5  \\
    \bottomrule
    \end{tabular}
  \label{tab:time}
\end{table*}

\begin{table*}[htbp]
  \centering
  \small
    \caption{$L_2$ regularization parameter configurations on continuous-time dynamics}
      \begin{tabular}{rllllll}
    \toprule
          &       & Grid  & Random & Power Law & Small World & Community \\
    \midrule
          & Adaptive-NDCN & 1e-3  & 1e-6  & 1e-3  & 1e-3  & 1e-3 \\
          \cmidrule{2-7}& GTS-NDCN & 1e-3  & 1e-6  & 1e-3  & 1e-3  & 1e-3 \\
\cmidrule{2-7}    Heat  & \textbf{SGODEv1} & 1e-3  & 1e-6  & 1e-3  & 1e-3  & 1e-3 \\
\cmidrule{2-7}    Diffusion & \textbf{SGODEv2} & 1e-3  & 1e-6  & 1e-3  & 1e-3  & 1e-3 \\
\cmidrule{2-7} & \textbf{SGODEv3} & 1e-3  & 1e-6  & 1e-3  & 1e-3  & 1e-3 \\
    \midrule
          & Adaptive-NDCN & 1e-3  & 1e-4  & 1e-5  & 1e-5  & 1e-4 \\
   \cmidrule{2-7}        & GTS-NDCN & 1e-3  & 1e-4  & 1e-5  & 1e-5  & 1e-4 \\
\cmidrule{2-7}    Mutualistic  & \textbf{SGODEv1} & 1e-3  & 1e-4  & 1e-4  & 1e-4  & 1e-4 \\
\cmidrule{2-7}    Interaction & \textbf{SGODEv2} & 1e-3  & 1e-4  & 1e-5  & 1e-5  & 1e-4 \\
\cmidrule{2-7} & \textbf{SGODEv3} & 1e-3  & 1e-4  & 1e-5  & 1e-5  & 1e-4 \\
    \midrule
          & Adaptive-NDCN & 1e-4  & 1e-4  & 1e-4  & 1e-4  & 1e-4 \\
  \cmidrule{2-7}         & GTS-NDCN & 1e-4  & 1e-4  & 1e-4  & 1e-4  & 1e-4 \\
\cmidrule{2-7}    Gene  & \textbf{SGODEv1} & 1e-4  & 1e-4  & 1e-4  & 1e-4  & 1e-4 \\
\cmidrule{2-7}    Regulation & \textbf{SGODEv2} & 1e-4  & 1e-4  & 1e-4  & 1e-4  & 1e-4 \\
\cmidrule{2-7}& \textbf{SGODEv3}  & 1e-4  & 1e-4  & 1e-4  & 1e-4  & 1e-4  \\
    \bottomrule
    \end{tabular}
  \label{tab:para_set1}
\end{table*}

\begin{table*}[htbp]
  \centering
  \small
    \caption{$L_2$ regularization parameter configurations on regular-sampled time series}
      \begin{tabular}{rllllll}
    \toprule
          &       & Grid  & Random & Power Law & Small World & Community \\
    \midrule
          & Adaptive-NDCN & 1e-3  & 1e-6  & 1e-3  & 1e-3  & 1e-3 \\
\cmidrule{2-7}    Heat  & \textbf{SGODEv1} & 1e-3  & 1e-6  & 1e-3  & 1e-3  & 1e-3 \\
\cmidrule{2-7}    Diffusion & \textbf{SGODEv2} & 1e-3  & 1e-6  & 1e-3  & 1e-3  & 1e-3 \\
\cmidrule{2-7} & \textbf{SGODEv3} & 1e-3  & 1e-6  & 1e-3  & 1e-3  & 1e-3 \\
    \midrule
          & Adaptive-NDCN & 1e-3  & 1e-4  & 1e-5  & 1e-5  & 1e-4 \\
\cmidrule{2-7}    Mutualistic  & \textbf{SGODEv1} & 1e-3  & 1e-4  & 1e-4  & 1e-4  & 1e-4 \\
\cmidrule{2-7}    Interaction & \textbf{SGODEv2} & 1e-3  & 1e-4  & 1e-5  & 1e-5  & 1e-4 \\
\cmidrule{2-7}& \textbf{SGODEv3} & 1e-5  & 1e-4  & 1e-4  & 1e-4  & 1e-4 \\
    \midrule
          & Adaptive-NDCN & 1e-4  & 1e-4  & 1e-4  & 1e-4  & 1e-4 \\
\cmidrule{2-7}    Gene  & \textbf{SGODEv1} & 1e-4  & 1e-4  & 1e-4  & 1e-4  & 1e-4 \\
\cmidrule{2-7}    Regulation & \textbf{SGODEv2} & 1e-4  & 1e-4  & 1e-4  & 1e-4  & 1e-4 \\
\cmidrule{2-7}& \textbf{SGODEv3}   & 1e-4  & 1e-4  & 1e-4  & 1e-4  & 1e-4 \\
    \bottomrule
    \end{tabular}
  \label{tab:para_set2}
\end{table*}

\subsection{Three synthetic dynamics}
We report the time of the synthetic dynamics in Table \ref{tab:time}. The dimension of the node embedding vector in SGODE is 10. The learning rate in the continuous dynamics and regularized sampling experiments is 0.01, and the weight decay parameters for irregular and equidistant sampling are reported in Tables \ref{tab:para_set1}  and \ref{tab:para_set2}. The parameters for NDCN and No-graph are based on the paper \cite{Zang2020Neural}. In addition, for GTS-NDCN, the regularization loss coefficient $\lambda=1$, $k=10$ for kNN, and the dimension of the node embedding vector is 10.

\subsection{Traffic datasets}
In METR-LA, we set the diffusion step to 2 and the node embedding size C to 10. The dimensionality of the hidden vector is 64. The learning rate is set to 0.005.

In PEMS-BAY, we set the diffusion step to 3 and the node embedding size C to 10. The dimensionality of the hidden vector is 128. The learning rate is set to 0.005.

In PeMSD4, for SGODE-NCDE, we set the number of K to 2 and the node
embedding size C to 10. The dimensionality of the hidden vector is 64. The learning rate is set to 0.001 and the weight decay is 0.001. For SGODE-RNN, we set the diffusion step to 2 and the node embedding size C to 8. The dimensionality of the hidden vector is 64. The learning rate is set to 0.005. 

In PeMSD8, for SGODE-NCDE, we set the number of K to 2 and the node
embedding size C to 2. The dimensionality of the hidden vector is 128. The learning rate is set to 0.001 and the weight decay is 0.001. For SGODE-RNN, we set the diffusion step to 2 and the node embedding size C to 6. The dimensionality of the hidden vector is 64. The learning rate is set to 0.005. 

\section{Three continuous-time dynamics}\label{3dynamics}
\label{3:Three dynamics}
\subsection{Details of three continuous-time dynamics}
We generate three continuous-time dynamics: heat diffusion dynamics, mutualistic interaction dynamics, and gene regulation dynamics on five graphs. We consider five graphs to simulate network dynamics \cite{Zang2020Neural}: 1) Grid network, where each node is connected with 8 neighbors; 2) Random network \cite{erdos1959random}; 3) Power-law network \cite{barabasi1999emergence}; 4) Small-world network \cite{watts1998collective}; 5) Community network \cite{fortunato2010community}. 

For continuous-time network dynamics, we randomly sample 120 graph snapshots with irregular time intervals. For the first 100 snapshots, we randomly select 80 as the training set and 20 as the testing set for interpolation prediction. The remaining 20 snapshots are used as the testing set for extrapolation prediction. For the regularly-sampled time series, we take the first 80 snapshots for training, and the remaining 20 snapshots for testing. Specifically, we use the Dormand-Prince method \cite{dormand2018numerical} to generate ground-truth values of each network dynamics with 400 nodes for all networks.

Let ${x_{i}(t)} \in \mathbb{R}^{d \times 1}$ be $d$ dimensional features of node $i$ at time $t$ and thus $\bm{X}(t)=[...,{x_{i}(t)},...]^\mathrm{T} \in \mathbb{R}^{n \times d}$. Three dynamics are described as follows:

\textit{1) Heat Diffusion Dynamics}. The heat diffusion dynamics on signed graphs can be modeled as $\frac{\mathrm{d} x_{i}\left ( t \right ) }{\mathrm{d} t}  =  -k_{i,j} \textstyle \sum_{j  =  1}^{n} [\bm{A}]_{ij} \left ( x_{i} - x_{j} \right)$. The heat flow is transferred from node $i$ to node $j$ through the connection from node $i$ to node $j$, where $k_{i,j}$ represents the heat diffusion coefficient \cite{luikov2012analytical,ma2008mining}.

\textit{2) Mutualistic Interaction Dynamics}.
The mutualistic interaction dynamics tracks the abundance ${x_i(t)}$ of species $i$ \cite{gao2016universal}, as depicted by $\frac{\mathrm{d} x_{i} \left ( t \right )}  {\mathrm{d} t}  =  b_{i} + x_{i} \left ( 1- \frac{x_{i}}{k_{i}} \right )\left (  \frac{x_{i}}{c_{i}} -1 \right ) + \textstyle \sum_{j =1}^{n} [\bm{A}]_{ij} \frac{x_{i}x_{j} }{ d_{i} + e_{i}x_{i} + h_{j}x_{j}}$. Considering that real systems are composed of numerous components, the mutualistic systems consist of incoming migration of $i$ at a rate $b_i$ from neighboring ecosystems, logistic growth with system carrying capacity $k_i$ \cite{zang2018power} and threshold of the Allee effect $c_i$.

\textit{3) Gene Regulatory Dynamics}. The gene regulatory dynamics  $\frac{\mathrm{d} x_{i} \left ( t \right ) } {\mathrm{d} t}  =  -b_{i} x_{i}^{f} + \textstyle \sum_{j =1}^{n} [\bm{A}]_{ij} \frac{x_{j}^{h}} { x_{j}^{h}+1}$ governed by the Michaelis-Menten equation \cite{alon2006introduction} describes degradation ($f=1$) or dimerization ($f=2$) in the first term and genetic activation tuned by the Hill coefficient $h$ in the second term \cite{gao2016universal}. 

\subsection{Results of extrapolation experiments for three continuous-time dynamics}
The extrapolation experiment results of irregular sampling and equidistant sampling are reported in Table \ref{tab:ex-irregular} and Table \ref{tab:ex-equal}.

\begin{table}[htbp]
  \centering
     \caption{MAPE of extrapolation of continuous-time network dynamics. I: heat diffusion dynamics; II: mutualistic interaction dynamics; III: gene regulatory dynamics; Com.: Community. Those in black font indicate the best performance. The underline corresponds to the second-ranked value.}
    \resizebox{\linewidth}{!}{
    \begin{tabular}{crrrrrr}
    \toprule
          &   Methods    & \multicolumn{1}{l}{Grid} & \multicolumn{1}{l}{Random} & \multicolumn{1}{l}{Power} & \multicolumn{1}{l}{Small} & \multicolumn{1}{l}{Com.} \\
    \midrule
    \multirow{5}[12]{*}{} & No-graph & 28.9±0.4 & 7.1±0.7 & 9.2±0.6 & 8.6±0.6 & 19.9±5.9 \\
          & NDCN  & \textbf{3.7±1.0} & 6.1±1.1 & 5.2±0.6  & \textbf{2.5±0.4} & 5.3±1.3 \\
     I     & Adaptive-NDCN & 9.3±2.5 & 20.4±4.9 & 12.6±2.9 & 13.9±4.7 & 12.9±4.1 \\
     & GTS-NDCN & 10.1±2.5 & 6.4±1.9 & 5.7±1.2 & 6.9±2.6 & 8.7±2.3 \\
      \cmidrule{2-7}    
      & \textbf{SGODEv1} & 6.3±1.1 & \textbf{2.5±0.7} & 4.3±0.9 & 5.7±1.3 & 5.6±1.3 \\
          & \textbf{SGODEv2} & 5.0±1.4 & 3.0±1.2 & \textbf{3.0±1.2} & \textbf{2.5±0.7} & \textbf{2.6±0.9} \\
          & \textbf{SGODEv3} & \underline{4.6±1.2} & \underline{2.9±1.8} & \underline{3.6±0.8} & \underline{4.3±1.4} & \underline{4.7±1.6} \\
    \midrule
    \multirow{5}[12]{*}{} & No-graph & 56.7±0.6 & 18.3±14.4 & 15.5±3.1 & 55.2±0.3 & 10.0±1.8 \\
          & NDCN  & 27.1±1.8 & 10.6±10.4 & 8.8±1.6 & \textbf{15.5±1.6} & \textbf{4.0±1.7} \\
     II    & Adaptive-NDCN & 56.3±0.2 & 5.8±5.4 & 14.6±9.1 & 34.6±6.2 & 13.3±3.4 \\
     & GTS-NDCN & \textbf{18.6±8.3} & 27.3±13.6 & 12.4±3.3 & 18.9±4.8 & 7.4±3.1 \\
        \cmidrule{2-7}  & \textbf{SGODEv1} & 24.8±2.4 & \underline{4.1±1.4} & 8.3±3.9 & \underline{15.9±2.9} & 10.0±1.7 \\
          & \textbf{SGODEv2} & 25.2±2.9 & \textbf{4.0±2.2} & \textbf{7.1±1.2} & 25.1±1.6 & \underline{5.7±1.8} \\
          & \textbf{SGODEv3} & \underline{19.1±2.9} & 4.7±1.3 & \underline{7.8±2.8} & 26.2±6.2 & 6.8±2.5 \\
    \midrule
    \multirow{5}[12]{*}{} & No-graph & 14.4±1.5 & 12.0±0.2 & 38.6±1.3 & 17.3±0.7 & 18.9±0.3 \\
         & NDCN  & 14.0±4.6 & 2.5±0.3 & 4.1±0.5 & \underline{3.8±0.5} & \textbf{2.2±0.7} \\
     III    & Adaptive-NDCN & 5.9±2.0 & \underline{1.6±0.9} & \underline{3.2±1.1} & 5.6±2.5 & 2.3±1.0 \\
     & GTS-NDCN & 5.3±1.8 & 3.0±0.4 & 8.7±1.1 & 6.8±1.7 & 4.1±1.4 \\
       \cmidrule{2-7}  & \textbf{SGODEv1} & \textbf{4.9±1.0} & 3.8±1.6 & \textbf{3.1±0.7} & 4.2±1.2 & 4.1±1.5 \\
          & \textbf{SGODEv2} & 8.8±1.4 & \textbf{1.5±1.0} & \underline{3.2±1.2} & \textbf{2.7±0.9} & \underline{2.1±1.4} \\
          & \textbf{SGODEv3} & \underline{4.8±1.7} & 2.8±1.9 & 3.7±1.3 & 4.8±2.7 & 3.3±1.6 \\
    \bottomrule
    \end{tabular}}
     \label{tab:ex-irregular}
\end{table}

\begin{table}[ht]
  \centering
     \caption{MAPE of extrapolation of regularly-sampled time series.}
    \resizebox{\linewidth}{!}{
    \begin{tabular}{crrrrrr}
    \toprule
          &   Methods    & \multicolumn{1}{l}{Grid} & \multicolumn{1}{l}{Random} & \multicolumn{1}{l}{Power} & \multicolumn{1}{l}{Small} & \multicolumn{1}{l}{Com.} \\
    \midrule
    \multirow{5}[12]{*}{} & LSTM-GNN & 14.8±2.8 & 22.0±5.8 & 7.9±8.0 & 11.0±1.3 & 11.4±2.6 \\
          & GRU-GNN & 11.8±2.3 & 14.4±3.1 & 8.1±1.2  & 9.0±1.8 & 7.9±1.3 \\
     I     & RNN-GNN & 19.1±5.1 & 25.1±5.1 & 11.6±4.7 & 20.4±3.1 & 12.9±2.9 \\
     & Adaptive-NDCN & \underline{10.7±2.1} & 16.5±6.4 & 18.7±6.2 & 11.8±5.3 & 15.8±3.4  \\
      \cmidrule{2-7}    & \textbf{SGODEv2} & 11.1±3.1 & \underline{7.6±3.9} & \underline{7.7±1.3} & \underline{8.9±2.3} & \underline{5.5±0.8}  \\
          & \textbf{SGODEv3} & \textbf{4.4±1.0} & \textbf{3.9±1.7} & \textbf{3.6±1.0} & \textbf{4.4±1.4} & \textbf{5.1±2.1}  \\
    \midrule
    \multirow{5}[12]{*}{} & LSTM-GNN & 51.3±3.1 & 27.1±19.7 & 24.1±13.4 & 58.7±6.1 &19.5±19.6 \\
          & GRU-GNN  & 51.7±2.4 &\textbf{1.7±4.0} & \underline{13.0±3.4} & 54.4±6.3 & 7.9±6.3 \\
     II    & RNN-GNN & 56.7±0.3 & 27.5±11.9 & 34.0±0.5 & 59.4±4.0 & 8.0±5.3 \\
     & Adaptive-NDCN & 53.7±5.4 & 12.0±5.8 & 17.1±8.1 & 38.6±12.4 & 5.0±1.4  \\
        \cmidrule{2-7}  & \textbf{SGODEv2} & \textbf{25.2±2.2} & 7.0±5.5 & \textbf{12.6±2.6} & \textbf{26.7±5.3} & \textbf{2.9±0.6}  \\
          & \textbf{SGODEv3} & \underline{25.4±2.2} & \underline{2.6±0.9} & 16.1±8.8 & \underline{27.7±2.8} & \underline{4.0±0.9} \\
    \midrule
    \multirow{5}[12]{*}{} & LSTM-GNN & 27.3±3.2 & 66.9±13.7 & 54.7±13.3 & 14.9±3.4 & 50.1±16.2 \\
         & GRU-GNN  &29.0±2.3 & 53.8±7.5 & 52.1±15.8 & 13.5±2.3 & 41.1±9.6 \\
     III    & RNN-GNN & 28.9±6.6 & 54.6±9.9 & 49.6±9.3 & 15.8±6.2 & 45.2±3.4 \\
     & Adaptive-NDCN & \underline{8.7±2.4} & \textbf{1.5±0.4} & \underline{4.0±0.9} & \underline{4.7±1.6} & 1.5±0.8  \\
       \cmidrule{2-7}  & \textbf{SGODEv2} & 10.5±3.2 & 4.4±2.1 & 5.1±1.9 & 11.1±3.4 & \underline{1.2±0.8} \\
          & \textbf{SGODEv3} & \textbf{7.0±4.1} & \underline{3.1±1.5} & \textbf{3.5±1.3} & \textbf{4.0±2.9} & \textbf{1.1±1.9}\\
    \bottomrule
    \end{tabular}}
     \label{tab:ex-equal}
\end{table}

\begin{figure}[htbp]
\centering
\subfloat[Fix embedding dim = 8 and vary $k$.] {\includegraphics[width=0.25\textwidth]{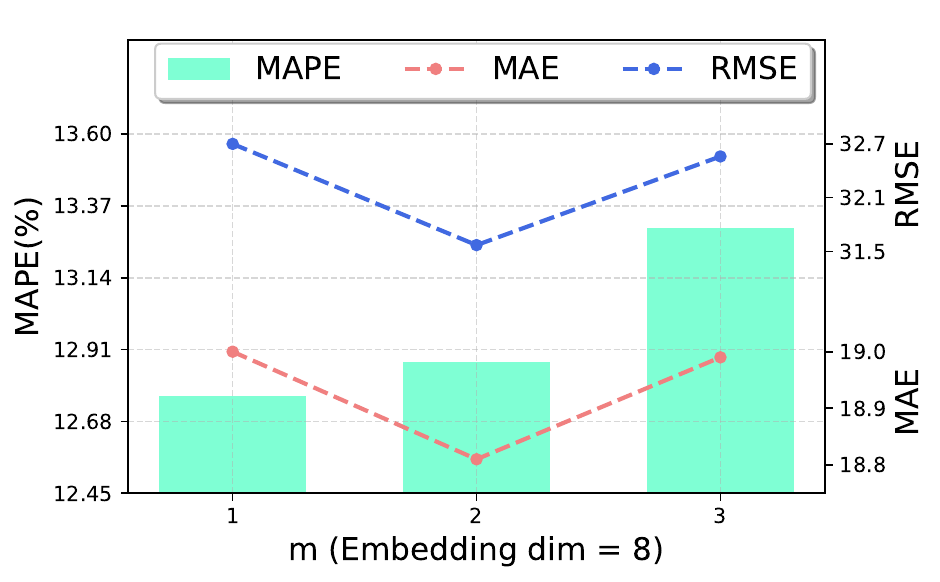}}
\subfloat[Fix $k$ = 2 and vary embedding dim.] {\includegraphics[width=0.25\textwidth]{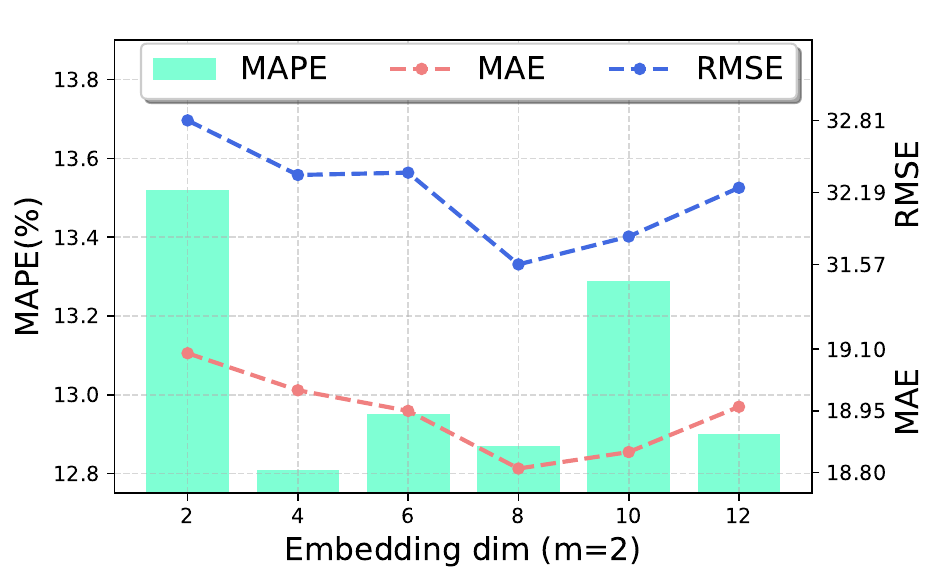}}
\caption{Sensitivity analysis of diffusion step and node embedding vectors on PeMSD4.}
\label{fig:para_PEMSD4}
\end{figure}

\begin{figure}[htbp]
\centering
\subfloat[Fix embedding dim = 6 and vary $k$.] {\includegraphics[width=0.25\textwidth]{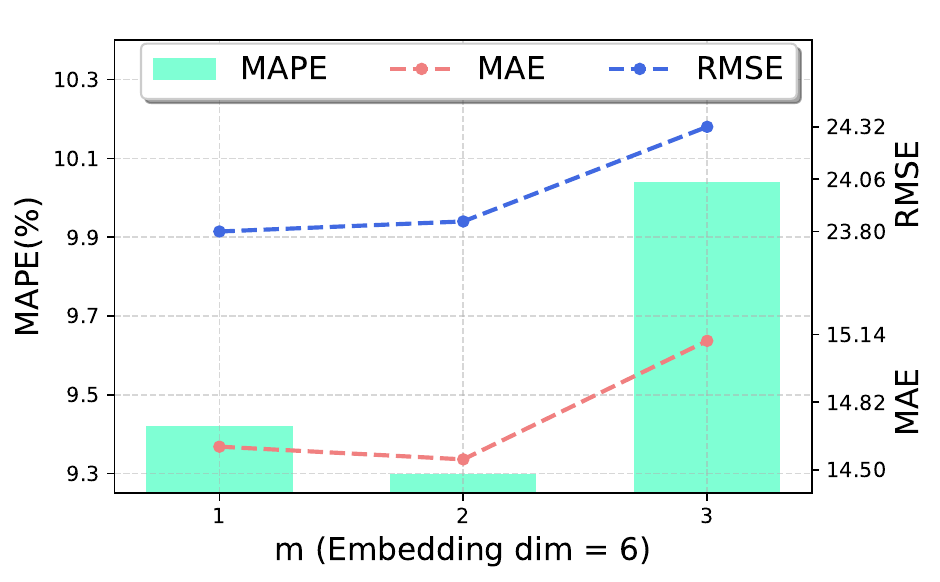}}
\subfloat[Fix $k$ = 2 and vary embedding dim.] {\includegraphics[width=0.25\textwidth]{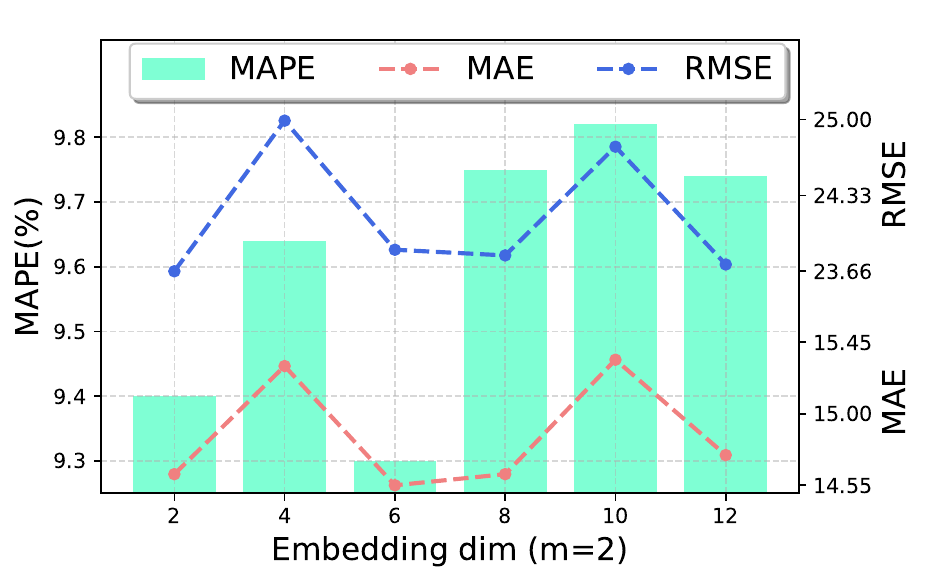}}
\caption{Sensitivity analysis of diffusion step and node embedding vectors on PeMSD8.}
\label{fig:para_PEMSD8}
\end{figure}

\section{Sensitivity analysis and prediction error at each horizon}\label{sensitivity}
Figs. \ref{fig:para_PEMSD4}, \ref{fig:para_PEMSD8}, and \ref{fig:para_PEMS-BAY} show the results by varying the diffusion step and the dimensionality of the node embedding vector respectively.

Fig.\ref{fig:RMSE} and Fig.\ref{fig:MAPE} show the RMSE and MAPE for each horizon not reported in our main paper.

\begin{figure}[htbp]
\centering
\subfloat[Fix embedding dim = 10 and vary $k$.] {\includegraphics[width=0.25\textwidth]{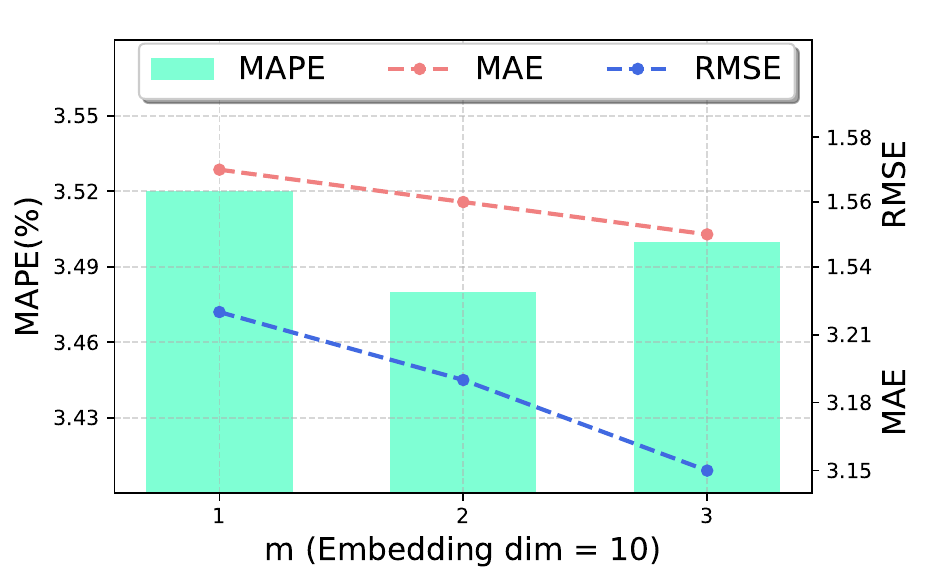}}
\subfloat[Fix $k$ = 3 and vary embedding dim.] {\includegraphics[width=0.25\textwidth]{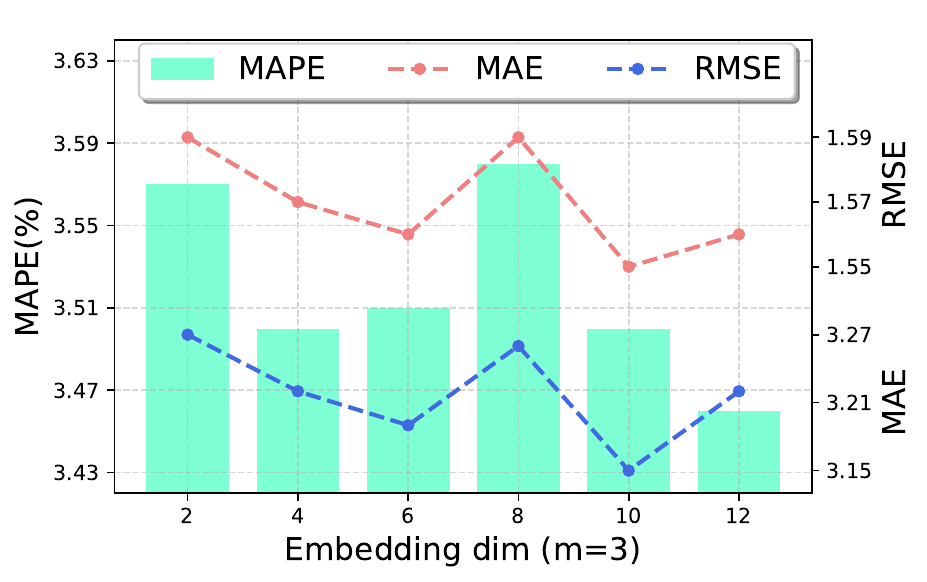}}
\caption{Sensitivity analysis of diffusion step and node embedding vectors on PEMS-BAY.}
\label{fig:para_PEMS-BAY}
\end{figure}

\begin{figure}[htbp]
\centering
\subfloat[RMSE on PeMSD4] {\includegraphics[width=0.25\textwidth]{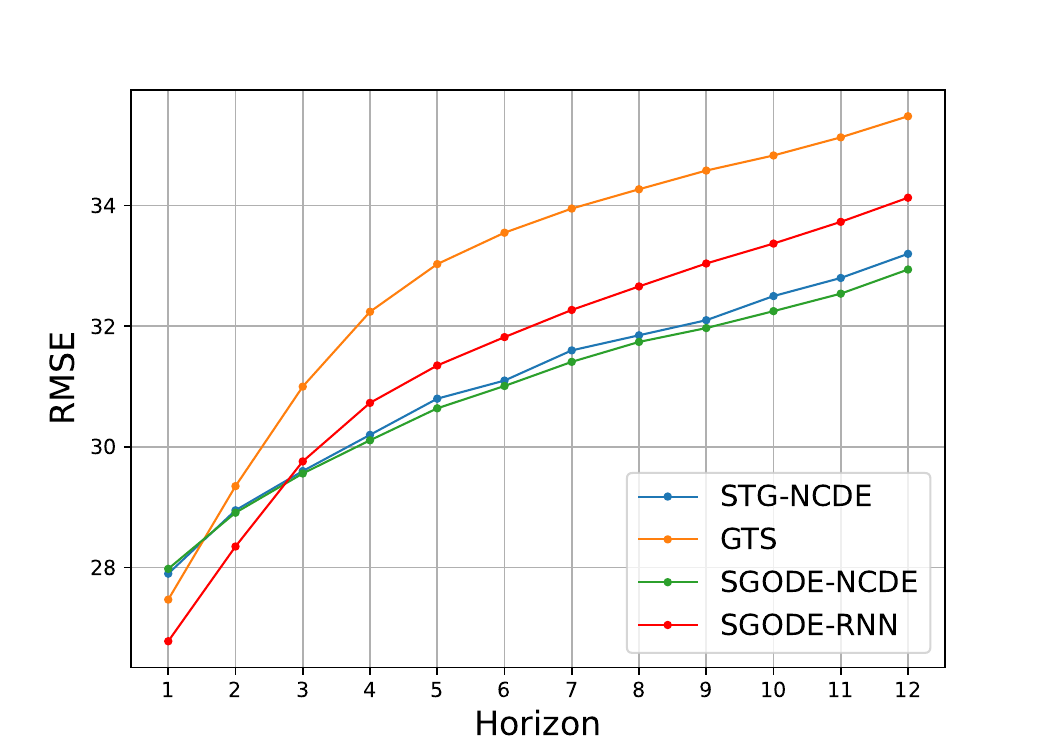}\label{fig:04RMSE}}
\subfloat[RMSE on PeMSD8] {\includegraphics[width=0.25\textwidth]{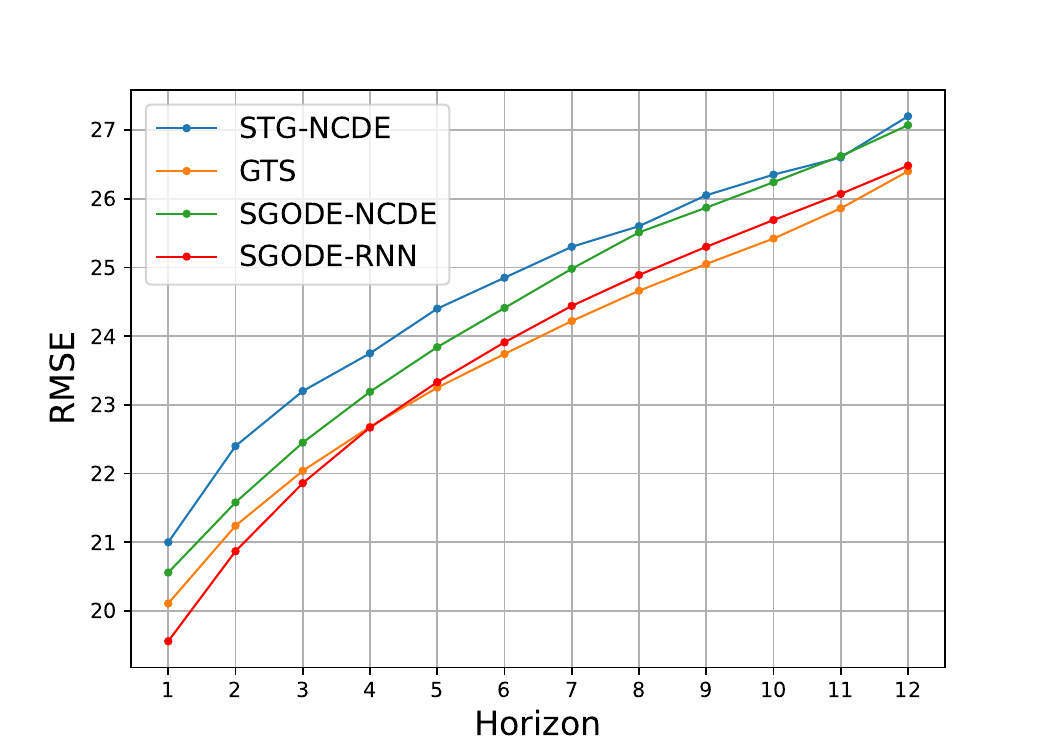}\label{fig:08RMSE}}
\caption{RMSE at each horizon.}
\label{fig:RMSE}
\end{figure}

\begin{figure}[htbp]
\centering
\subfloat[MAPE on PeMSD4] {\includegraphics[width=0.25\textwidth]{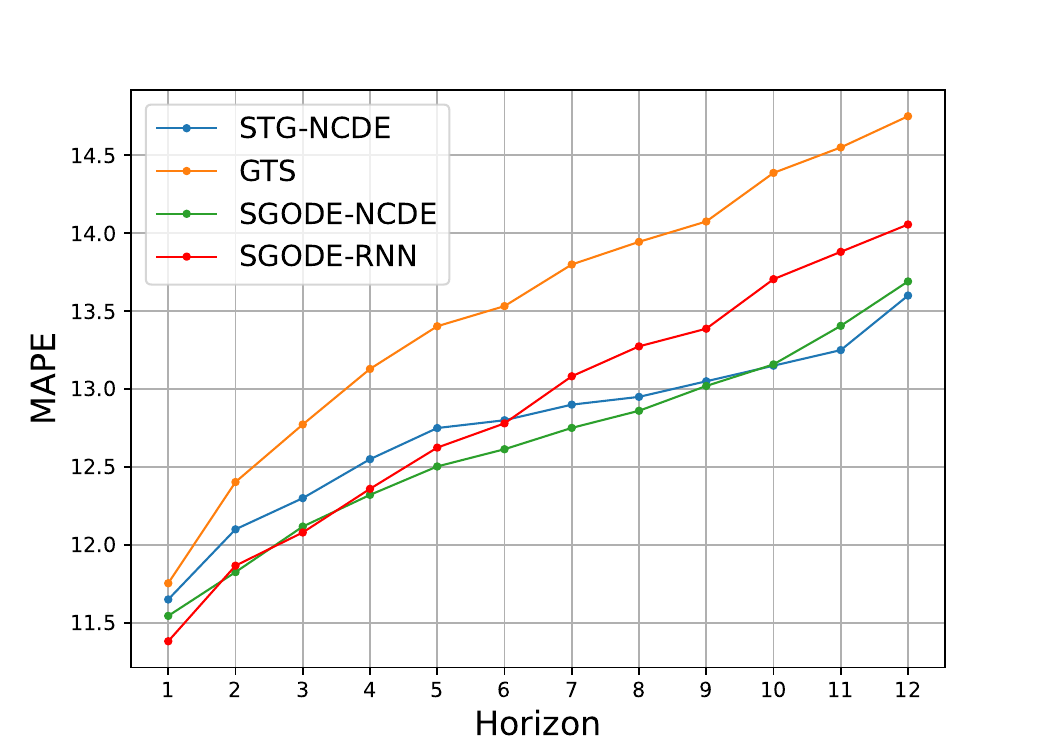}\label{fig:04MAPE}}
\subfloat[MAPE on PeMSD8] {\includegraphics[width=0.25\textwidth]{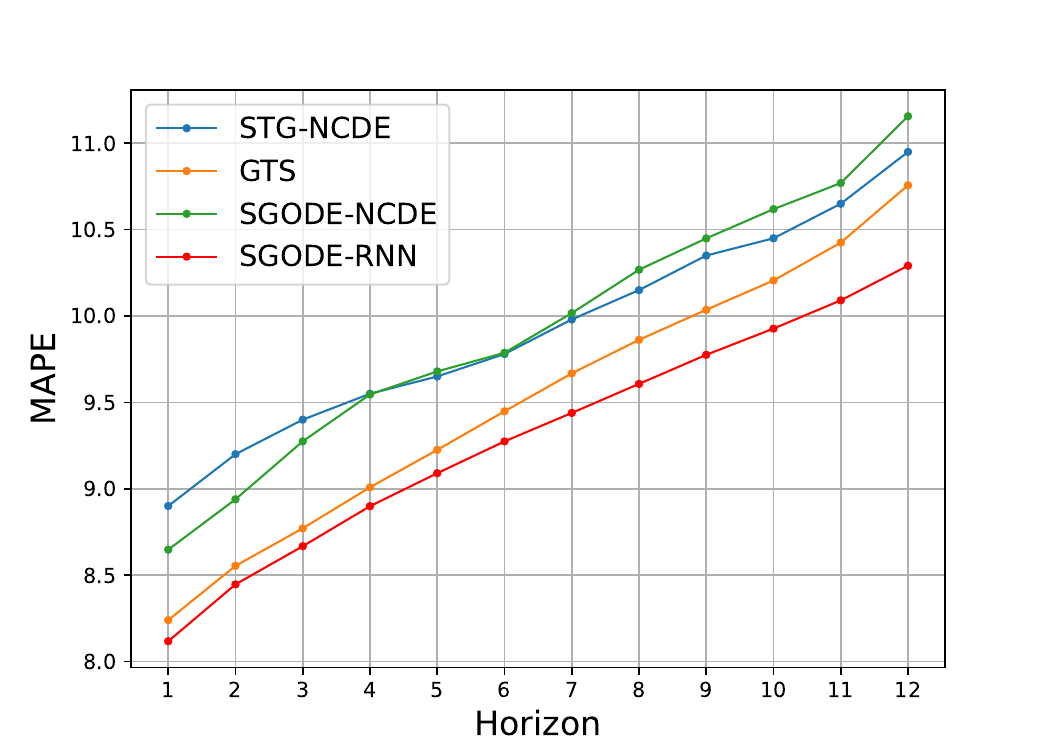}\label{fig:08MAPE}}
\caption{MAPE at each horizon.}
\label{fig:MAPE}
\end{figure}

\section{Metrics}
\label{s2}
Suppose $x = x_1, \cdots , x_n$ represents the ground truth, $\hat{x} = \hat{x}_1, \cdots, \hat{x}_n$ represents the predicted values, and $\Omega$ denotes the indices of observed samples, the metrics are defined as follows.

Mean Absolute Percentage Error (MAPE):

\begin{equation}
    \mathrm{MAPE} = \frac{1}{\Omega}\sum_{i \in \Omega} \left|\frac{x_i-\hat{x}_i}{x_i}\right|
\end{equation}

Mean Squared Error (MSE):
\begin{equation}
    \mathrm{MSE} = \frac{1}{\Omega}\sum_{i \in \Omega} \left|({x_i-\hat{x}_i})^2\right|
\end{equation}

\section{Different Results of GWNET}

We noticed it is acknowledged in existing literature like [$*$] \footnote{\footnotesize{[$*$] Exploring Progress in Multivariate Time Series Forecasting: Comprehensive Benchmarking and Heterogeneity Analysis.}\label{p1}} that significant differences in the results can be observed for the same baseline across different papers(especially BasicTS) due to three main factors.
When it comes to our results compared to BasicTS, we point out two main factors: 1) BasicTS incorporates external temporal features into the raw data, such as attributes for time of the day and day of the week, while we do not; 2) BasicTS uses masked MAE as the loss function, whereas we employ native MAE. 
Despite these differences, we still achieve competitive results.

\begin{table}[htbp]
\vskip -0.15in
  \centering
  \resizebox{\linewidth}{!}{
    \begin{tabular}{ccccccc}
    \toprule
    \multirow{2}[4]{*}{Model} & \multicolumn{3}{c}{PeMSD4} & \multicolumn{3}{c}{PeMSD8} \\
\cmidrule{2-7}          & MAE   & RMSE  & MAPE  & MAE   & RMSE  & MAPE \\
    \midrule
    DCRNN$^1$ & 19.66 & 31.18 & 13.45\% & 15.23 & 24.17 & 10.21\% \\
    GraphWaveNet$^1$ & \textbf{18.80} &  \textbf{30.14} & 13.19\% & \underline{14.67} & \textbf{23.55} & \underline{9.46\%} \\
    SGODE-NCDE & 19.06 & \underline{30.96} & \textbf{12.65\%} & 15.34 & 24.44 & 9.92\% \\
    SGODE-DCRNN & \underline{18.81} & 31.57 & \underline{12.87\%} & \textbf{14.55} & \underline{23.85} & \textbf{9.30\%} \\
    \bottomrule
    \end{tabular}}
    \vskip -0.15in
  \label{tab:pems1}
\end{table}

\section{Further Explanation of Performance Variations}
In Table below, we provide a summary of the recommended version to use for different scenarios. The difference between \textbf{v1} and \textbf{v2}, \textbf{v3} lies in the parameter count of $K$, where \textbf{v1} has $O(n^2)$ parameters, and \textbf{v2} and \textbf{v3} have $O(nd)$ parameters. In most cases, $d$ is significantly smaller than $n$. The primary distinction between \textbf{v2} and \textbf{v3} in $K$ is the presence of zero elements. Clearly, when the number of nodes $n$ is small, \textbf{v1} is the most suitable choice. However, as $n$ increases, \textbf{v2} and \textbf{v3} become better options. In many cases, there are no relationships between numerous nodes in the underlying graph, making \textbf{v3} the preferred choice for general scenarios. We conduct experiments with SGODE-NCDE on PeMSD4, and the MAE result with \textbf{v2} is 19.66, while with \textbf{v3}, it improved to 19.06. The synthetic dataset length $L < n$, so the results are sensitive to the parameter count, which is why \textbf{v2} performs slightly better than \textbf{v3} in more challenging irregular sampling experiments(except for the grid graph).

\begin{table}[htbp]
\vskip -0.15in
  \centering
  \resizebox{\linewidth}{!}{
    \begin{tabular}{cccccc}
    \toprule
   D:dataset & small D & small D& large D& large D\\
   S: regular sampling  & (RS) & (not RS)&(RS) & (not RS)\\

    \midrule
    small graph & \textbf{SGODEv1} & \textbf{SGODEv1}&   \textbf{SGODEv3}&   \textbf{SGODEv3}\\
    grid graph & \textbf{SGODEv3} & \textbf{SGODEv3}&   \textbf{SGODEv3}&   \textbf{SGODEv3} \\
    other graphs &  \textbf{SGODEv2} &   \textbf{SGODEv3}&   \textbf{SGODEv3}&   \textbf{SGODEv3}\\
    \bottomrule
    \end{tabular}}
    \vskip -0.15in
  \label{tab:select}
\end{table}

\end{document}